\documentclass[lettersize,journal]{IEEEtran}
\usepackage{amsmath,amsfonts}
\usepackage{algorithmic}
\usepackage{algorithm}
\usepackage{array}
\usepackage{subfigure}
\usepackage[caption=false,font=normalsize,labelfont=sf,textfont=sf]{subfig}
\usepackage{textcomp}
\usepackage{stfloats}
\usepackage{url}
\usepackage{verbatim}
\usepackage{graphicx}
\usepackage{cite}
\usepackage{adjustbox}
\usepackage{multirow}
\usepackage{makecell}
\usepackage{soul}
\newcolumntype{C}[1]{>{\centering\arraybackslash}m{#1}}

\usepackage{xcolor}
\usepackage{colortbl}

\definecolor{Gray}{gray}{0.9}
\definecolor{DarkGray}{gray}{0.8}
\definecolor{LightGray}{gray}{0.97}

\newcolumntype{g}{>{\columncolor{Gray}}c}
\newcolumntype{h}{>{\columncolor{LightGray}}c}
\newcolumntype{d}{>{\columncolor{DarkGray}}c}

\usepackage[capitalize]{cleveref}
\crefname{section}{Sec.}{Secs.}
\Crefname{section}{Section}{Sections}
\Crefname{table}{Table}{Tables}
\crefname{table}{Tab.}{Tabs.}

\begin{document}

\title{Ancilia: Scalable Intelligent Video Surveillance for the Artificial Intelligence of Things}

\author{Armin Danesh Pazho\IEEEauthorrefmark{1},~\IEEEmembership{Student Member,~IEEE,}
        Christopher Neff\IEEEauthorrefmark{1},~\IEEEmembership{Student Member,~IEEE,}
        Ghazal Alinezhad Noghre,~\IEEEmembership{Student Member,~IEEE,}
        Babak Rahimi Ardabili,~\IEEEmembership{Student Member,~IEEE,}
        Shanle Yao,
        Mohammadreza Baharani,~\IEEEmembership{ Member,~IEEE,}
        Hamed Tabkhi,~\IEEEmembership{Member,~IEEE}
\thanks{The authors are with the Electrical and Computer Engineering Department, The University of North Carolina at Charlotte, Charlotte,
	NC, 28223 USA.\\
	\{adaneshp, cneff1, galinezh, brahimia, mbaharan, htabkhiv\}@uncc.edu\\
	\IEEEauthorrefmark{1} Corresponding authors have equal contribution.}}

\markboth{IEEE INTERNET OF THINGS JOURNAL}%
{Danesh Pazho \MakeLowercase{\textit{et al.}}: Title: To Be Determined}


\maketitle

\begin{abstract}
With the advancement of vision-based artificial intelligence, the proliferation of the Internet of Things connected cameras, and the increasing societal need for rapid and equitable security, the demand for accurate real-time intelligent surveillance has never been higher. This article presents Ancilia, an end-to-end scalable, intelligent video surveillance system for the Artificial Intelligence of Things. Ancilia brings state-of-the-art artificial intelligence to real-world surveillance applications while respecting ethical concerns and performing high-level cognitive tasks in real-time. Ancilia aims to revolutionize the surveillance landscape, to bring more effective, intelligent, and equitable security to the field, resulting in safer and more secure communities without requiring people to compromise their right to privacy.
\end{abstract}


\section{Introduction}\label{sec:Introdution}
There is a growing need for effective and efficient surveillance technologies that can be deployed to protect our cities, people, and infrastructure. For example, in Itaewon, South Korea, a holiday celebration left over 150 dead due to severe overcrowding, with many blaming the tragedy on careless government oversight \cite{itaewon}. In Moore County, North Carolina, directed attacks against two power substations left over 45,000 residents without power for days as technicians rushed to restore power and authorities struggled to find the source of the attacks \cite{NCPower}. With enough forewarning through smart video surveillance, they could have been prevented. 

With the recent emergence of the Artificial Intelligence of Things (AIoT), some surveillance solution providers have started adding basic forms of artificial intelligence to their systems. However, their methods are still naive and unable to enhance security in a truly meaningful way \cite{feldstein2019global}. This is because, while a lot of research is conducted on tasks that would benefit surveillance systems, most works focus on algorithmic improvements in a lab environment instead of paying attention to factors that are prevalent in real-world scenarios \cite{AnomalySurvey, AnomalySurvey2}. Most research focuses on a single algorithm and how to tweak it to get the best possible results on readily available datasets that often do not reflect a real surveillance environment. Few works explore how different algorithms affect the performance of other downstream algorithms in multi-algorithm systems. Few still explore the effects of noise (both data derived and the system produced) in end-to-end accuracy. Beyond this, real-world intelligent surveillance necessitates real-time performance. The cognitive abilities of advanced artificial intelligence are only helpful if they can be provided to security personnel quickly enough to take appropriate action before it is too late.

\begin{figure}[h!] 
	\centering
	\includegraphics[width=1\linewidth, trim= 18 12 15 15,clip]{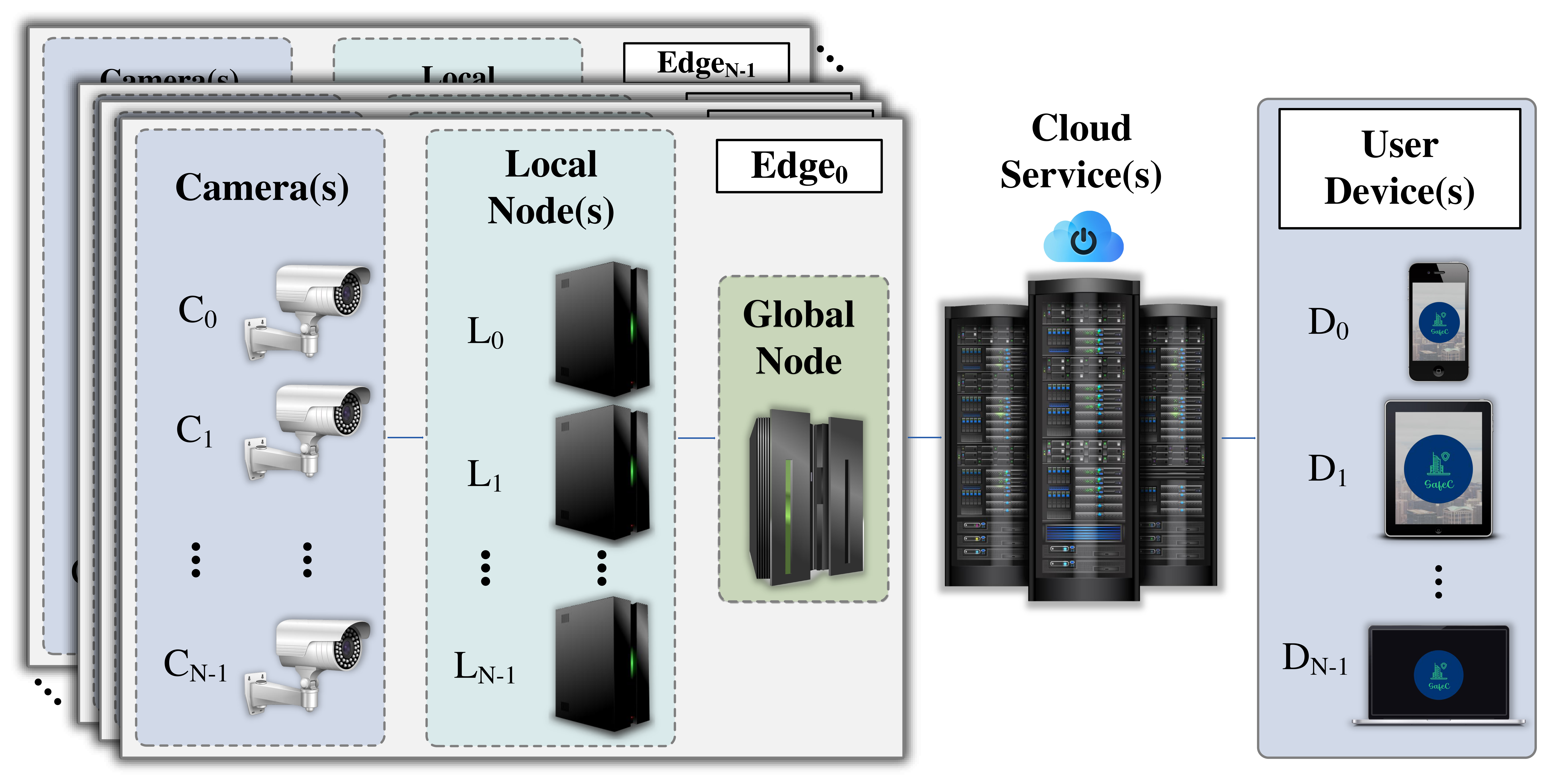}
	\captionsetup{justification=centering}
	\caption{Conceptual overview of Ancilia.}
	\label{fig:intro}
\end{figure}

In this article, we present Ancilia, the first end-to-end scalable, intelligent video surveillance system able to perform high-level cognitive tasks in real-time while achieving state-of-the-art results. Ancilia takes advantage of the prevalence of cameras in the Internet of Things (IoT) and uses localized servers to integrate with existing IoT camera ecosystems, facilitating processing on the edge. Current IoT methods often use cloud computing, which can introduce latency and privacy concerns, or they require custom sensors with high processing power. Ancilia offers a solution to utilize existing IoT sensors, minimizing the need for expensive infrastructure upgrades and reliance on cloud processing. Ancilia is device agnostic; As long as video from the camera can be accessed, Ancilia can provide intelligence. Shown in \cref{fig:intro}, Ancilia exists within three logical and physical segments: the edge, the cloud, and user devices. The edge uses a plethora of advanced artificial intelligence algorithms processing data received from cameras to facilitate intelligent security. Using a single workstation to perform edge processing, Ancilia can monitor up to 4 cameras in real-time at 30 FPS, or up to 8 cameras at 15 FPS, in scenarios with both medium and heavy crowd density. Ancilia performs high-level cognitive tasks (i.e. action recognition, anomaly detection) with $\sim 1\%$ deviation in accuracy from current State-of-the-Art (SotA).

Ancilia is designed from the ground up to respect the privacy of the people and communities being surveilled. Ancilia does not store any personally identifiable information in any databases and does not make use of invasive artificial intelligence techniques such as facial recognition or gait detection. Ancilia strictly provides pose and locational information for high-level tasks (i.e. action recognition, anomaly detection), as opposed to identity information, which is common. Ancilla looks at what a person is doing, not who they are. This allows Ancilia to act as a buffer to help remove biases based on race, ethnicity, gender, age, and socio-economic factors, which can lead to a reduction in the unnecessary conflict between authorities and marginalized communities that has become increasingly problematic. After data is processed on edge and sent to the cloud for communication and service management with user devices. A mobile app allows user devices to receive data from the cloud, including alerts when potential security concerns arise.

In summary, this article has the following contributions:

\begin{itemize}
    \item We present Ancilia, the first end-to-end scalable real-world intelligent video surveillance system capable of performing high-level cognitive tasks in real-time while achieving SotA accuracy. 
    \item We analyze the ethical concerns of intelligent video surveillance, both from a privacy and fairness perspective, and illustrate how Ancilia's design is purpose-built to address them.
    \item We perform an end-to-end empirical evaluation of Ancilia using two high-level cognitive tasks directly related to intelligent surveillance, action recognition, and anomaly detection, investigating the trade-off in accuracy required to achieve real-time performance.
    \item We perform an exhaustive system-level evaluation of Ancilia's real-time performance and scalability across different classes of hardware and increasing scenario intensities, displaying how Ancilia is able to meet real-time intelligent security needs in different contexts.
\end{itemize}

\section{Related Work}\label{sec:RelatedWork}

There has been a plethora of research regarding the use of artificial intelligence for video surveillance \cite{survey1, survey2, survey3, AnomalySurvey}. \cite{IndustryPervasive} proposes the use of region proposal based optical flow to suppress background noise and a bidirectional Bayesian state transition strategy to model motion uncertainty to enhance spatio-temporal feature representations for the detection of salient objects in surveillance videos. \cite{MaskTracking} proposes the use of a person detector, tracking algorithm, and mask classifier for tracking pedestrians through surveillance video streams. 

In \cite{AnomalySurvey}, it is determined that in order to address the latency concerns of real-time video surveillance, a shift towards edge computing is needed. Nikouei et al. \cite{preiSENSE,iSENSE3,iSENSE} explore the feasibility of using low-power edge devices to perform object detection and tracking in surveillance scenarios. They argue that in worst case 5 FPS is high enough throughput for tracking humans in surveillance applications, and as such computation can be pushed to the edge. However, their results show that even light weight convolutional neural networks can prove problematic for low-power devices, often reducing throughput below the 5 FPS threshold. \cite{REVAMP2T} proposes a system using low-power embedded GPUs to perform detection, tracking, path prediction, pose estimation, and multi-camera re-identification in a surveillance environment, while placing a focus on real-time execution and the privacy of tracked pedestrians. \cite{TX2MTMC} proposes a similar system, focusing solely on object detection, tracking, and multi-camera re-identification to increase throughput. \cite{INES} proposes using a combination of lightweight object detection models on the edge and more computationally expensive models in the cloud, splitting computation between the two to provide real-time video surveillance in a construction site environment. \cite{IoTParkingOccupancy} proposes the use of background detection, vehicle detection, and kalman filter \cite{kalman} based tracking for parking lot surveillance and determining lot occupancy. \cite{Peeking} proposes a system that uses object detection, person tracking, scene segmentation, and joint trajectory and activity prediction for pedestrians in a surveillance setting.

The future of intelligent surveillance is heading towards systems able to perform high-level cognitive tasks. A recent survey focusing on real-world video surveillance \cite{AnomalySurvey} asserts that while the domain of video surveillance is comprised of understanding stationary object, vehicles, individuals, and crowds, the ability to determine when anomalous events occur is paramount for intelligent surveillance systems. Other research has supported this assertion \cite{survey1}. \cite{BayesNP} utilizes the Infinite Hidden Markov Model and Bayesian Nonparametric Factor Analysis to find patterns in video streams and detect abnormal events. \cite{ISTL} proposes active learning and fuzzy aggregation to learn what constitutes an anomaly continually over time, adapting the scenarios not seen in standard datasets. \cite{MallBehaviors} proposes a system to detect suspicious behaviors in a mall surveillance setting, using lightweight algorithms such as segmentation, blob fusion, and kalman filter based tracking \cite{kalman}. AnomalyNet \cite{AnomalyNet} is a recently proposed recurrent neutral network with adaptive iterative hard-thresholding and long short-term memory that works directly off pixel information to eliminate background noise, capture motion, and learn sparse representation and dictionary to perform anomaly detection in video surveillance.

\section{Ethical Concerns}\label{sec:Ethics}
Video surveillance has always been associated with social and ethical concerns, whether in traditional form or more recent intelligent formats. Respecting citizens' privacy and autonomy while improving public safety and security are the most well-known and enduring ethical issues in this context \cite{10.1145/3313831.3376347, nissenbaum2004privacy, appenzeller2020ethical,PrivacySurveillance}. Developing a successful smart video surveillance solution that addresses the public safety problem and engages the community up to a certain level is only possible by considering these concerns. 

There is rising attention among scholars to the issue of incorporating privacy concerns at the design level, referred to as "privacy by design"  \cite{hartzog2018privacyos}. The source of discrimination and privacy violation in many data-driven and AI-based systems, such as Smart video surveillance technology, is using Personal Identifiable Information (PII)\cite{daubert2015view, speicher2018potential}.
Using PII, such as actual footage of people's daily activities at any stage of the technology, can increase the risk of privacy violation. There is a long-lasting debate on the ethical challenges of using facial recognition technologies in different sectors and how using this technology can result in privacy violation\cite{raji2020saving, martinez2019important,introna2010facial, selinger2021ethics}. 

The approaches used to perform high-level cognitive tasks in intelligent video surveillance, such as action recognition and anomaly detection, can be grouped into two distinct categories based on the data used \cite{angelini20192d}. The first category directly utilizes pixel data. A common example is facial recognition \cite{guo2011complexity}, where algorithms look at images of people's faces to identify them. These algorithms can perform well with sufficient historical data, but are often seen as intrusive and increase the risk of identifying personal demographic information \cite{introna2010facial}. The second category only leverages processed information, such as pose data in the case of Ancilia, which tends to de-identify personal demographic information \cite{ozyer2021human}. This is not a complete removal of PII, as some works have been able to identify individuals purely by gait \cite{liao2020model} or silhouette \cite{ccreid}, but it significantly reduces the risk to privacy compared to pixel-based approaches.

Similarly, avoiding facial recognition technologies does not guarantee the system is entirely privacy persevering. Storing images of pedestrians is another source of ethical violation. From the discrimination perspective, using any form of PII can contribute to the issue of marginalization in policing systems\cite{https://doi.org/10.5281/zenodo.4050457}. Therefore, an essential step in designing a non-discriminatory system is to ensure the system is not dependent on PII. This requires a specific approach toward the design of such technology in the choice of algorithm, the type of data used, and the storing of such data. 

Ancilia addresses this by not storing any PII or sending any PII across the network. Such data is destroyed after it is used. Ancilia utilizes pose-based methods for all high-level cognitive tasks, severely limiting the amount and quality of PII used by such algorithms. This allows such processing without any potential for gender, ethnicity, or class-based discrimination. As such, Ancilia is able to address many of the privacy concerns regarding intelligent video surveillance while also addressing the ethical issue of discrimination.
\section{Ancilia Algorithmic Framework}\label{sec:Algorithm}

\begin{figure*}[t] 
	\centering
	\includegraphics[width=1\linewidth, trim= 16 15 20 18,clip]{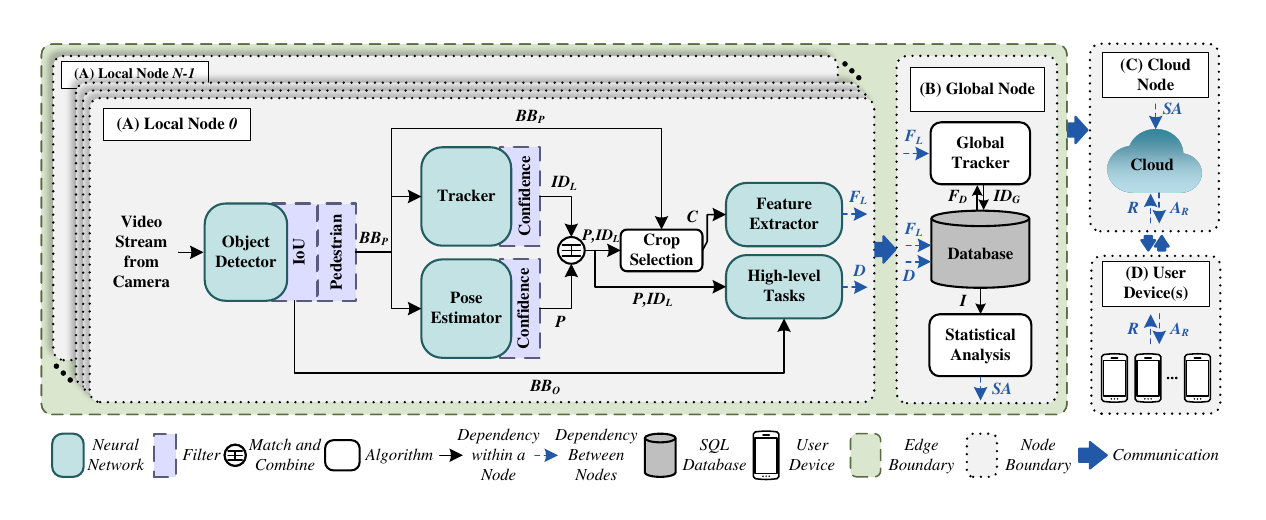}
	\captionsetup{justification=centering}
	\caption{Ancilia algorithmic details. $N$ local nodes are connected to a single global node on the edge. The final analyses are transferred to the cloud node to feed the application on the user device. Multiple edges may be connected to the could, though this figure only shows one edge for clarity. $BB_P$, $BB_O$, $ID_L$, $P$, $C$, $F_L$, $D$, $F_D$, $ID_G$, $I$, $SA$, $R$, and $A_R$ refer to bounding boxes for pedestrians, bounding boxes of objects, local identities, poses, person crops that passed selection, features from the local node, data from the downstream tasks, features from the database, global identities, information from the database, completed statistical analysis, requests from users, and requested attributes respectively.}
 \vspace{-15pt}
	\label{fig:algo}
\end{figure*}

The algorithmic core of Ancilia is separated into two conceptual systems: the local nodes containing the algorithmic pipeline of each camera and the global node that handles all processing that requires understanding of multiple camera perspectives. These two systems make up the algorithmic core of Ancilia and are the basis on which all higher understanding is achieved. A visual representation of this algorithmic core can be seen in \cref{fig:algo}.

\subsection{Single Camera Vision Pipeline}
As seen in \cref{fig:algo}, the local algorithmic pipeline starts when an image is extracted from the camera. The image is first run through an object detector to locate people, vehicles, animals, and other important objects in the scene. This is important not only because it acts as the basis for the rest of the algorithmic pipeline but also because it can be used for basic situational awareness. Sometimes, just the presence of a certain object in a scene is noteworthy, like a person in an unauthorized location, a bag left unattended, or the presence of a firearm. Ancilia uses YOLOv5 \cite{YOLOv5} for this purpose (however, it can be any detector). Please note that many objects of interest are not included in the default weights provided by the YOLOv5 authors. However, other works have trained the architecture for classes such as firearms \cite{warsi2019gun, narejo2021weapon, garg2021intelligent}, and custom weights can always be trained to match the target application. The locational coordinates of persons are sent to a tracker, where tracklets are created, matching each person with their previous detections in prior images. Ancilia utilizes the version of ByteTrack \cite{bytetrack} without frame similarity. In this configuration, ByteTrack does not perform feature extraction, which results in a notable reduction in computation. As shown in \cite{bytetrack}, locational similarity is sufficient for single camera tracking. The tracking allows for understanding how a person moves throughout a scene, which is vital for many surveillance applications. It also allows Ancilia to understand which poses belong to which persons over time, which is vital for many high-level tasks that provide much-needed situational awareness. Image crops of the people detected in the image are also sent to a human pose estimator, where two-dimensional pose skeletons are created. Ancilia uses HRNet \cite{hrnet} for extracting 2D skeletons. Using pose data for higher-level tasks has two major benefits over simply using raw pixel data. First, pose data is of much lower dimensionality than pixel data, making it much less computationally expensive and allowing the Ancilia to function in real-time. Second, pose data helps us remove the appearance-based PII information inherent in pixel data, making it harder for high-level tasks to form unintended biases based on ethnicity, gender, age, or other identity-based metrics. Works such as \cite{an2018improving, hasan2020multi} try to identify subjects based on their poses, in a line of work called Gait Recognition, but as discussed in \cref{sec:Ethics}, pose-based approaches are shown to be more privacy preserving compared to their alternatives.

\begin{figure*}[t] 
	\centering
	\includegraphics[width=1\linewidth, trim= 16 15 20 16,clip]{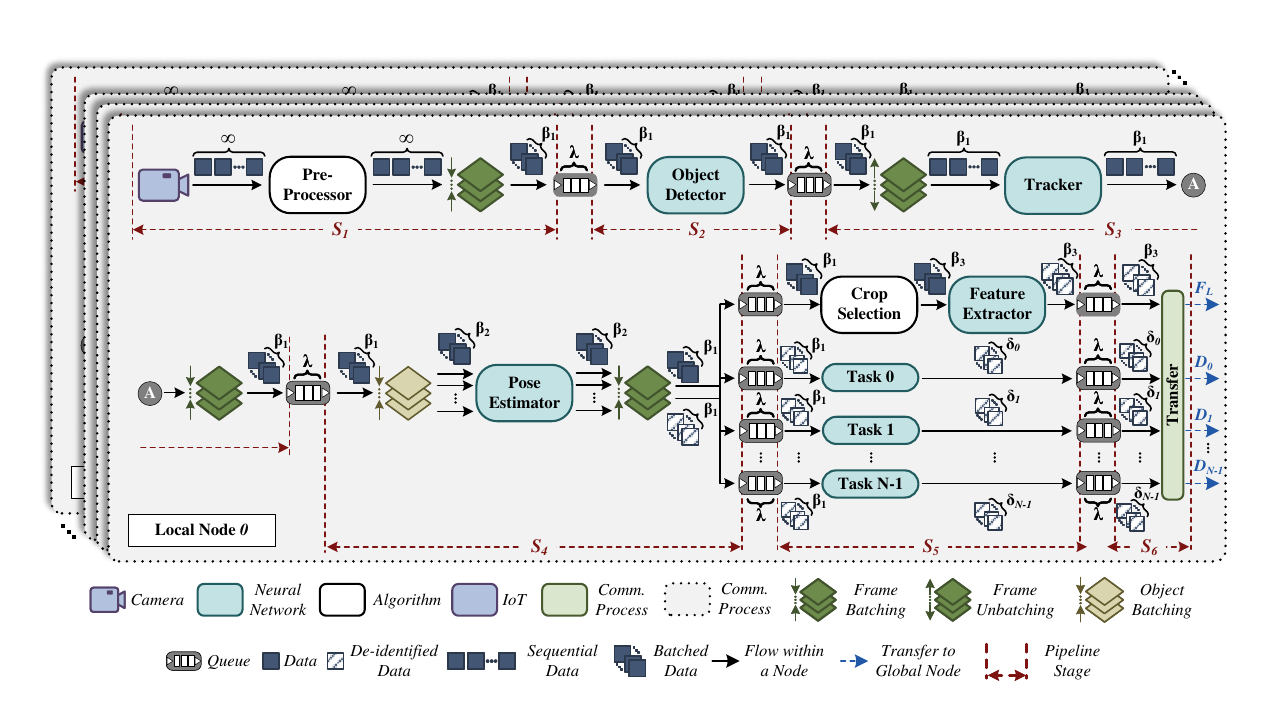}
	\captionsetup{justification=centering}
	\caption{A detailed view of system design in Ancilia's local nodes. $\beta$ and $\delta$ refer to different batch sizes. $\lambda$ refers to the queue size. $F_L$ and $D$ represent local features and data received from downstream tasks respectively.}
\vspace{-15pt}	
 \label{fig:system}
\end{figure*}

\subsection{Multi-Camera Person Re-identification}
While the tracker tracks people within a single camera, locational information cannot accurately re-identify a person across multiple cameras. For this, the same person crops that are sent to the human pose estimator are also sent to a person re-identification feature extractor, where an abstract feature representation is created for each person. Only one feature representation is created for each person during a single batch, and only when the quality of the representation can be assured, as poor quality representations are detrimental to accurate multi-camera person re-identification. Ancilia uses a feature representation filtering algorithm to verify two qualities for person crops. First, a person crop must contain a high-quality view of the person. To this end, the filter algorithm uses the 2D pose skeleton and verifies that at least 9 keypoints were detected with at least 60\% confidence. The filter algorithm looks at the overlap (i.e. Intersection of Union) of the bounding boxes generated by the object detector. An individual's bound box must have an Intersection over Union (IoU) of no more than 0.1 with any other person. If those two conditions are met, the person crop is determined to be of high enough quality to produce an adequate feature representation. If more than one crop is deemed suitable for a single person during a 15 frame window, the one with the most confident pose is selected. The features created by the feature extractor are sent to the global node for multi-camera person re-identification. Ancilia uses OSNet \cite{osnet} to extract feature representations.

\subsection{higher Level Tasks}
To help preserve privacy from a system perspective, sensitive information is kept on the local machine by executing all high-level tasks on the local node. These tasks have access to the object, tracking, and pose data generated in the previous steps. Since the decision of which high-level tasks are needed is highly application dependent, we do not consider these tasks to be part of Ancilia's algorithmic core, but instead an extension to be customized based on intended use. In this paper, we use action recognition and anomaly detection as two common examples of high-level tasks that are highly relevant to intelligent surveillance. For action recognition, we choose PoseConv3D \cite{duan2022revisiting} and CTR-CGN \cite{chen2021channel}, two state-of-the-art networks that can utilize the 2D human pose skeletons provided by Ancilia. For anomaly detection, we use GEPC \cite{markovitz2020graph} and MPED-RNN \cite{morais2019learning}, which are based on 2D human pose skeletons.

There are many works that use pixel-based methods for these tasks that achieve superior accuracies than SotA posed-based methods, such as I3D \cite{I3D}, MVIT \cite{MVIT} and Stargazer \cite{stargazer}. Argus \cite{argus} is a good example of a system that employs pixel-level information, with a subsequent evaluation conducted on a real-world surveillance dataset called Meva \cite{meva}. However, due to the privacy benefits discussed in \cref{sec:Ethics} and the computational benefits of using lower-dimensional pose data, we have opted to stick with pose-based methods for Ancilia.

\section{System Design}\label{sec:System}

Beyond the algorithmic design, Ancilia can be analyzed from a system-level design and implementation perspective. The local node in particular has a complex system design, as seen in \cref{fig:system}. The global node and cloud are much simpler, as shown in \cref{fig:algo}.

\subsection{Parallelism}
A key design objective of Ancilia is to achieve higher efficiency by balancing throughput and latency. Ancilia uses pipelining to take advantage of process parallelism, dividing tasks into six separate stages of a pipeline system ($S1, S2, ..., S6$). Each stage is implemented as a separate process, which executes concurrently with other processes as soon as it receives its required input. These stages communicate with each other using queues to utilize memory resources better and enable fast inter-process communication. While pipelining is a well-known technique for optimization, the overhead associated with its implementation means a balance needs to be found. \Cref{fig:system} shows a detailed view of the system design on the local node. Each pipeline stage is separated by a queue with a size limit of $\lambda$ elements, preventing any potential overflow from uneven execution speed between pipeline stages. By default, Ancilia uses a $\lambda$ value of 1. As is common, Ancilia offloads highly parallel tasks that rely on neural networks (i.e. object detection, pose estimation, feature extraction, and many high-level tasks) to Graphics Processing Units (GPUs) for execution.

\subsection{Data Batching}
Batching is another technique Ancilia implements to better utilize hardware resources. Generally, batching is able to greatly increase the throughput of a system at the cost of end-to-end latency. However, many high-level tasks (e.g. action recognition, anomaly detection) require multiple video frames worth of input data (often called a window) before they can start processing, so the latency that would be incurred by batching input frames is already inherent in these high-level task, as long as the frame batch and high-level task window are of the same size. In other words, if a high-level task needs $X$ number of frames before it can start processing, having a batch size of $X$ frames will ensure the task gets all the frames it needs simultaneously, incurring no additional latency for the task. If the window size of the high-level task is larger than the batch size multiple batches will be needed to be processed to receive output from the high-level task. Further, as frame batching ultimately increases the throughput, the end-to-end latency is decreased when compared processing each frame sequentially. While object detection works on entire frames, all other neural networks in Ancilia work off individual objects. These objects are batched together before being input to the network, greatly increasing hardware utilization. There can be multiple object batches within a single frame batch, based on how many of the relevant objects are detected in the video.

\subsection{Local Node}
\subsubsection{$S_1$ - Preprocessing}
Once the local node receives the video stream from the camera, the preprocessor is responsible for all basic image processing necessary before sending the frames through the algorithmic core. That includes any necessary resizing, frame dropping, and/or color channel reordering. Frame dropping is a dynamic machanism that ensures the framerate fed to the pipeline matches the throughput of the pipeline. For example, if the frame source (i.e. camera) produces 60 FPS, but Ancilia can only run at 30 FPS, only every second frame from the source will be passed through preprocessing. After preprocessing, frames are batched in sequential segements of size $\beta_1$. Ancilia sets $\beta_1=15$. This is done to balance throughput and latency, as discussed in \cref{sec:batch_size}, as well as to more closely match the window size of the high-level tasks, requiring only two batches to complete before these tasks can produce an output. This is also suitable because most modern security and IoT cameras record video at either 30 or 60 FPS. 

\subsubsection{$S_2$ - Object Detection}
The batched frames are sent to the object detector, which outputs a list of objects with class labels and bounding box coordinates. Bounding boxes for pedestrians are sent to the tracker, while bounding boxes for other objects are passed through the system for use in high-level tasks and statistical analysis. A crop of each pedestrian from the original frame is passed through to the pose estimator at later stages.

\subsubsection{$S_3$ - Tracking}
At the tracker, bounding boxes for pedestrians are unbatched to fit the tracker's sequential operation. The tracker groups the pedestrians and either matches them with previously seen pedestrians or assigns them a unique local ID. Afterwards, the pedestrians are once again batched by frame, back to the original batch size of $\beta_1=15$ frames, and sent to the pose estimator.

\subsubsection{$S_4$ - Pose Estimation}
At the pose estimator, the object batching is performed on the person crops, with a batch size of $\beta_2=32$. These batches are fed to the pose estimator, which outputs human pose skeletons for each person crop. Then the pedestrian bounding boxes, person crops, local IDs, and human pose skeletons are once again batched by frame and combined with the object bounding boxes from the object detector. Select data (pedestrian bounding boxes, person crops, local IDs, and pose skeletons) is sent to crop selection and feature extraction, while the de-identified data (pedestrian bounding boxes, object bounding boxes, local IDs, and pose skeletons) is sent to each high-level task as per their request.

\subsubsection{$S_5$ - Feature Extraction and High-level Tasks}
Before feature extraction, crop selection filters out low-quality person crops based on bounding box overlap and keypoint confidence, as described in \cref{sec:Algorithm}. By default, crops with an IoU higher than 0.1 or with 9 or more keypoints with confidence below 60\% are discarded. These thresholds can be adjusted to best suit the target application. Out of the remaining crops only a single crop with the highest overall keypoint confidence for each person is selected. The remaining crops are batched, with a dynamic size of $\beta_3$ based on the number of persons in the scene. Feature extractor receives the batch of $\beta_3$ crops. Once features are extracted, they are sent for transfer to the global node.
Each high-level task receives data at the granularity of a frame batch with size $\beta_1$, and sends data to the global node at the granularity that task operates at ($\delta_0, \delta_1, ..., \delta_n$). Only de-identified data is sent to the high-level tasks, keeping in line with the ethical concerns mentioned in \cref{sec:Ethics} and \cref{sec:Algorithm}. Each high-level task has its own process and works in parallel with other tasks as well as with crop selection and feature extraction in stage 5 of the pipeline.

\subsubsection{$S_6$ - Transfer}
Communication is completely decoupled from the pipeline, so once the data is sent, the local node pipeline continues to function as normal without needing a response from the global node. Importantly, no identifiable information is ever sent to the global node, keeping in line with the privacy and ethical concerns mentioned in \cref{sec:Ethics}.

\subsection{Global Node}
All received data is stored in a relational database on the global node. The matching algorithm described in \cref{sec:Algorithm} compares the received features with existing features in the database over the period $T$ and assigns a global ID based on the results. The default value for $T$ is set to 1 hour, but this should be changed to suit the needs of the application. An assortment of algorithms performs statistical analysis using the relational database, as detailed in \cref{sec:Algorithm}. The analysis is transmitted to the cloud node using APIs provided by the cloud service provider. By default, Ancilia uses Amazon Web Services, but this can be altered based on user/application needs. The cloud (e.g. Amazon Web Services (AWS)) receives analyzed data from the global node.

\section{Experimental Results}\label{sec:Results}

\subsection{Algorithmic Core}

The algorithmic core of Ancilia consist of multiple algorithms, each of which works off of data generated by the previous algorithms. As these algorithms leverage imperfect neural networks, they generate noise that accumulates through the system. To understand the source of this noise, we must first look at the accuracy of each of these core algorithms in isolation. \Cref{tab:baseline} shows the accuracies of the algorithmic core's four main tasks: object detection, pedestrian tracking, human pose estimation, and person re-identification. The table also shows the accuracies of the top SotA models in each task. These SotA methods are not suitable for intelligent surveillance applications, as their excessive computation and vast parameters make real-time execution impossible, but the comparison allows us to see the maximum potential allowable by current research and the accuracy loss incurred to keep Ancilia performing in real-time.

\begin{table}[]
\renewcommand{\arraystretch}{1.1}
\centering
\caption{Accuracy of Ancilia's Algorithmic Core networks in isolation. SotA Algorithms represent the highest performance currently achievable when computation and latency are not a concern.}
\label{tab:baseline}
\resizebox{\columnwidth}{!}{%
\begin{tabular}{cccc}
\rowcolor{DarkGray}
\multicolumn{1}{c|}{Task}                                                       & \multicolumn{1}{c|}{Method}                                    & \multicolumn{1}{c|}{Performance}  & Dataset                  \\ \hline \hline
\rowcolor{Gray}
\multicolumn{4}{c}{Ancilia's Algorithmic Core}                                                                                                                                                                  \\ \hline \hline
\multicolumn{1}{c|}{\begin{tabular}[c]{@{}c@{}}Object\\ Detection\end{tabular}} & \multicolumn{1}{c|}{YOLOv5 \cite{YOLOv5}}                      & \multicolumn{1}{c|}{49.0 (mAP)}   & COCO \cite{COCO}         \\ \hline
\multicolumn{1}{c|}{Tracking}                                                   & \multicolumn{1}{c|}{ByteTrack \cite{bytetrack}}                & \multicolumn{1}{c|}{77.8 (MOTA)}  & MOT20 \cite{MOT20}       \\ \hline
\multicolumn{1}{c|}{\begin{tabular}[c]{@{}c@{}}Pose\\ Estimation\end{tabular}}  & \multicolumn{1}{c|}{HRNet \cite{hrnet}}                        & \multicolumn{1}{c|}{75.1 (AP)}    & COCO \cite{COCO}         \\ \hline
\multicolumn{1}{c|}{\begin{tabular}[c]{@{}c@{}}Person\\ ReID\end{tabular}}      & \multicolumn{1}{c|}{OSNet \cite{osnet}}                        & \multicolumn{1}{c|}{88.6 (Top-1)} & DukeMTMC \cite{DukeMTMC} \\ \hline \hline
\rowcolor{Gray}
\multicolumn{4}{c}{State-of-the-Art Algorithms}                                                                                                                                                                 \\ \hline \hline
\multicolumn{1}{c|}{\begin{tabular}[c]{@{}c@{}}Object\\ Detection\end{tabular}} & \multicolumn{1}{c|}{Internimage \cite{internimage}}            & \multicolumn{1}{c|}{65.0 (mAP)}   & COCO \cite{COCO}         \\ \hline
\multicolumn{1}{c|}{Tracking}                                                   & \multicolumn{1}{c|}{SOTMOT \cite{SOTMOT}}                      & \multicolumn{1}{c|}{77.9 (MOTA)}  & MOT20 \cite{MOT20}       \\ \hline
\multicolumn{1}{c|}{\begin{tabular}[c]{@{}c@{}}Pose\\ Estimation\end{tabular}}  & \multicolumn{1}{c|}{ViTPose \cite{vitpose}}                    & \multicolumn{1}{c|}{81.1 (AP)}    & COCO \cite{COCO}         \\ \hline
\multicolumn{1}{c|}{\begin{tabular}[c]{@{}c@{}}Person\\ ReID\end{tabular}}      & \multicolumn{1}{c|}{Centeroids-ReID \cite{EffectiveCentroids}} & \multicolumn{1}{c|}{95.6 (Top-1)} & DukeMTMC \cite{DukeMTMC}
\end{tabular}%
}
\vspace{-18pt}
\end{table}

Object detection sees the biggest hit to accuracy, with a 16\% drop from SotA. This is intuitive, as YOLOv5 \cite{YOLOv5} is not only the largest model in the algorithmic core, but also the only one that operates on the raw camera stream. So while larger models are available and would be able to produce higher accuracy, even a slight increase in model size or computation would result in a noticeable decrease in throughput. Human pose estimation sees a decrease in accuracy for a similar reason, though much smaller in scale at only 6\%. While HRNet \cite{hrnet} is not run on the raw camera stream, it is run individually for each person detected by the object detector. As such, maintaining a small model size is preferable. Person re-identification sees a slightly larger drop in accuracy than human pose estimation at 7\%. While this is partly due to using a lightweight model, OSNet \cite{osnet}, the SotA model for person reID is also lightweight. However, the SotA uses a centroid based retrieval method not suitable for pen-set reID, of which most surveillance scenarios are. Pedestrian tracking sees almost no drop in accuracy, approximately 0.1\%. This stems from the comparative ease of tracking pedestrians in a single camera, where a simple, lightweight algorithm like ByteTrack \cite{bytetrack} see almost no performance difference from the top of the line SotA approaches.

\subsection{High-level Tasks} \label{sec:highlevel}

To better understand how the noise generated by the algorithmic core effects overall performance, and thus how well Ancilia performs in the realm of real-world intelligent surveillance, we examine the performance of two high-level cognitive surveillance tasks when running on Ancilia. For Ancilia to be a benefit to intelligent surveillance tasks, we must ensure that excess false alarms or missed positive events do not occur. To assess this, we choose action recognition and anomaly detection, as these tasks can utilize the human pose information generated by the algorithmic core, resulting in faster and less biased inference. Since both these methods utilize temporal batches of human poses for each individual, these experiments will directly reflect the quality of the object detection, tracking, re-identification, and pose estimation data generated by Ancilia.

\subsubsection{High-level Task - Action Recognition} \label{sec:action}

We select two state-of-the-art action recognition models, PoseConv3d \cite{duan2022revisiting} and CTR-GCN \cite{chen2021channel}, and train them using data generated with Ancilia. For each model, we train and test with full (30 FPS) and half (15 FPS) throughput on NTU60-XSub \cite{liu2019ntu}. 
Both models use a window size of 30 and are trained for 24 epochs using Stochastic Gradient Descent (SGD) with a momentum of 0.9 and Cosine Annealing scheduling. PoseConv3d and CTR-GCN have weight decay of $3e^{-4}$ and $5e^{-4}$ and an initial learning rate of 0.4 and 0.2, respectively.

\begin{table}[h]
\renewcommand{\arraystretch}{1.2}
\centering
\vspace{-5pt}
\caption{Top-1 and Top-5 accuracies on NTU60-XSub \cite{liu2019ntu} in full and half throughput modes for PoseConv3D \cite{duan2022revisiting} and CTR-GCN \cite{chen2021channel}.}
\vspace{-6pt}
\label{tab:action}
\begin{adjustbox}{max width=1\columnwidth,center}
\begin{tabular}{c||c|c||cc}
\rowcolor{DarkGray}
Model & Data & FPS & Top-1 (\%) & Top-5 (\%) \\ \hline \hline
\multirow{4}{*}{ PoseConv3D \cite{duan2022revisiting}} & \multirow{2}{*}{\cite{duan2022PYSKL}} & 15 & 91.96 & 99.47 \\ \cline{3-5} 
 &  & 30 & 92.76 & 99.57 \\ \cline{2-5} 
 & \multirow{2}{*}{Ours} & 15 & 88.79 & 98.82 \\ \cline{3-5} 
 &  & 30 & 91.99 & 99.28 \\ \hline \hline
\multirow{4}{*}{CTR-GCN \cite{chen2021channel}} & \multirow{2}{*}{\cite{duan2022PYSKL}} & 15 & 86.36 & 98.46 \\ \cline{3-5} 
 &  & 30 & 83.07 & 98.26 \\ \cline{2-5} 
 & \multirow{2}{*}{Ours} & 15 & 81.58 & 97.52 \\ \cline{3-5} 
 &  & 30 & 80.44 & 97.2
\end{tabular}
\end{adjustbox}
\vspace{-5pt}
\end{table}

The results of these experiments can be seen in \cref{tab:action}. We report the Top-1 and Top-5 accuracy and compare the results using data generated by Ancilia to the original data available through the PYSKL toolbox \cite{duan2022PYSKL}. We can see that Ancilia is able to provide data of comparable quality to the original; action recognition as a high-level task in Ancilia sees around 1\% drop in accuracy compared to the original data using PoseConv3D \cite{duan2022revisiting} at full throughput, and around 3\% at half throughput. Using CTR-GCN \cite{chen2021channel}, Ancilia sees a 2.5\% drop at full throughput and a 4.8\% drop at half throughput, compared to the original data. From this we can infer that PoseConv3D is more robust to noise than CTR-GCN, however both performed reasonably well with data generated from Ancilia, demonstrating its efficacy for intelligent surveillance applications.


\begin{figure*}[htb]
  \begin{tabular}{ccc}
    \subfigure[Server A]{\includegraphics[width=0.32\textwidth]{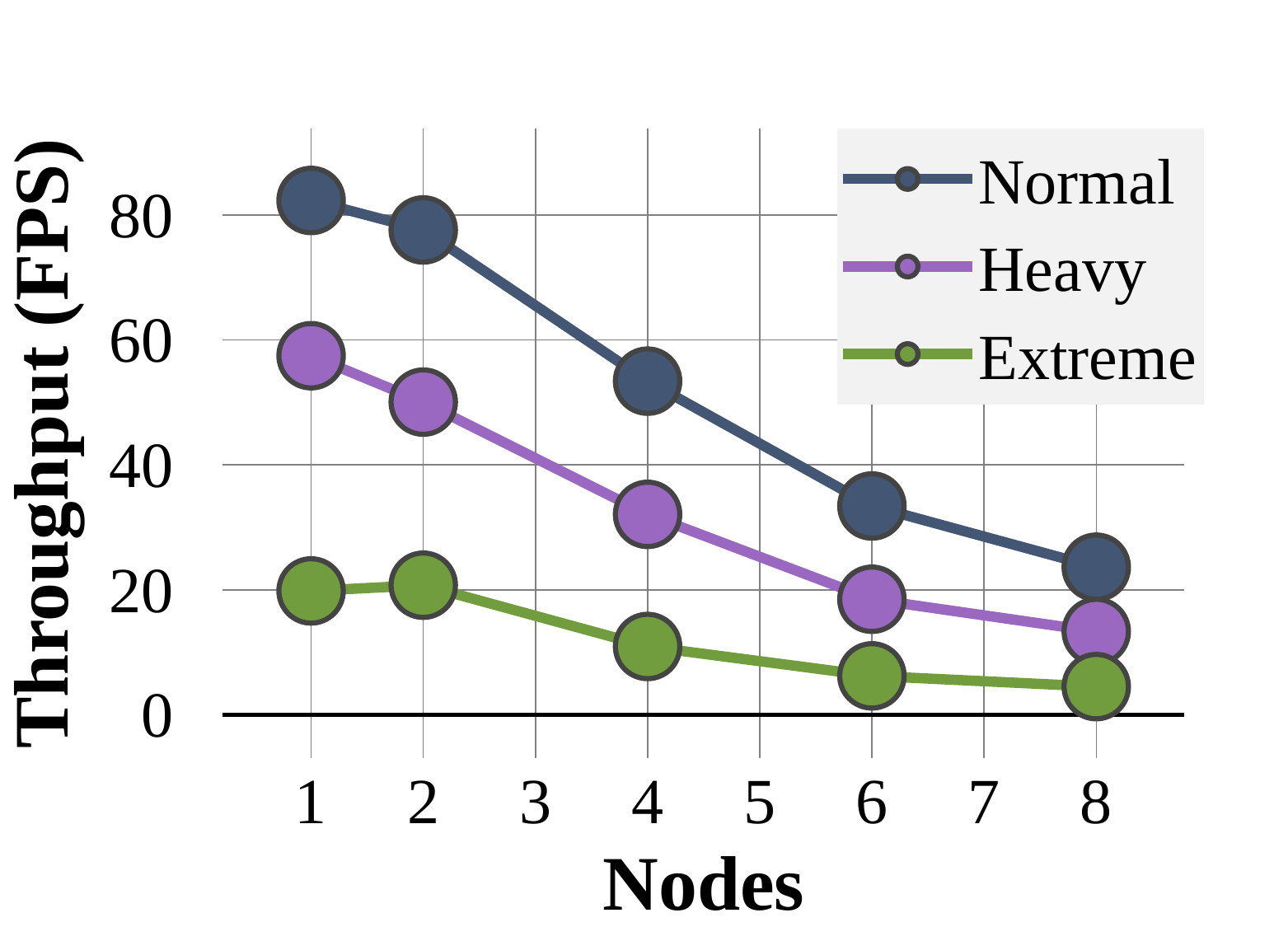}} &  
    \subfigure[Server B]{\includegraphics[width=0.32\textwidth]{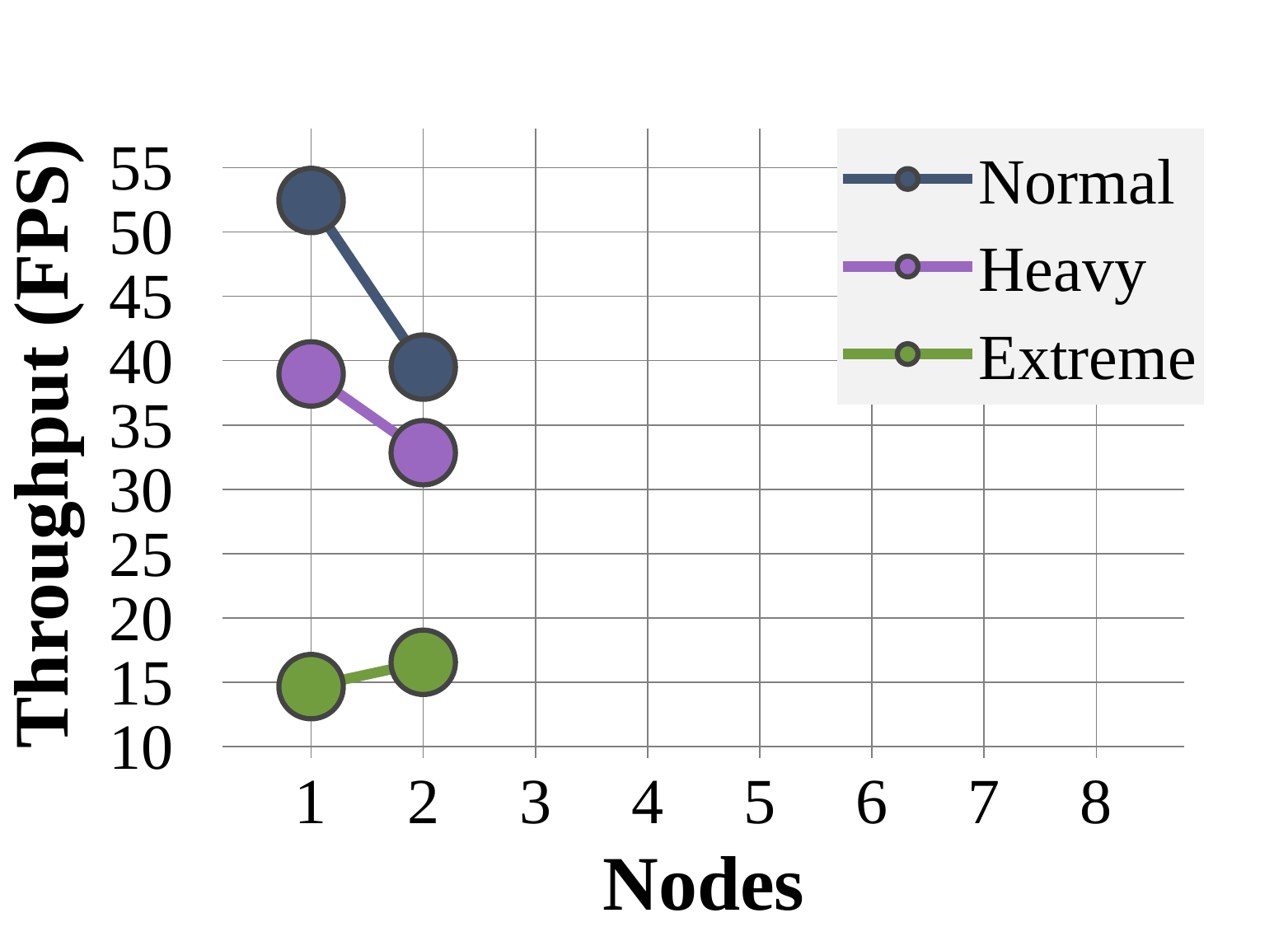}} &  
    \subfigure[Workstation]{\includegraphics[width=0.32\textwidth]{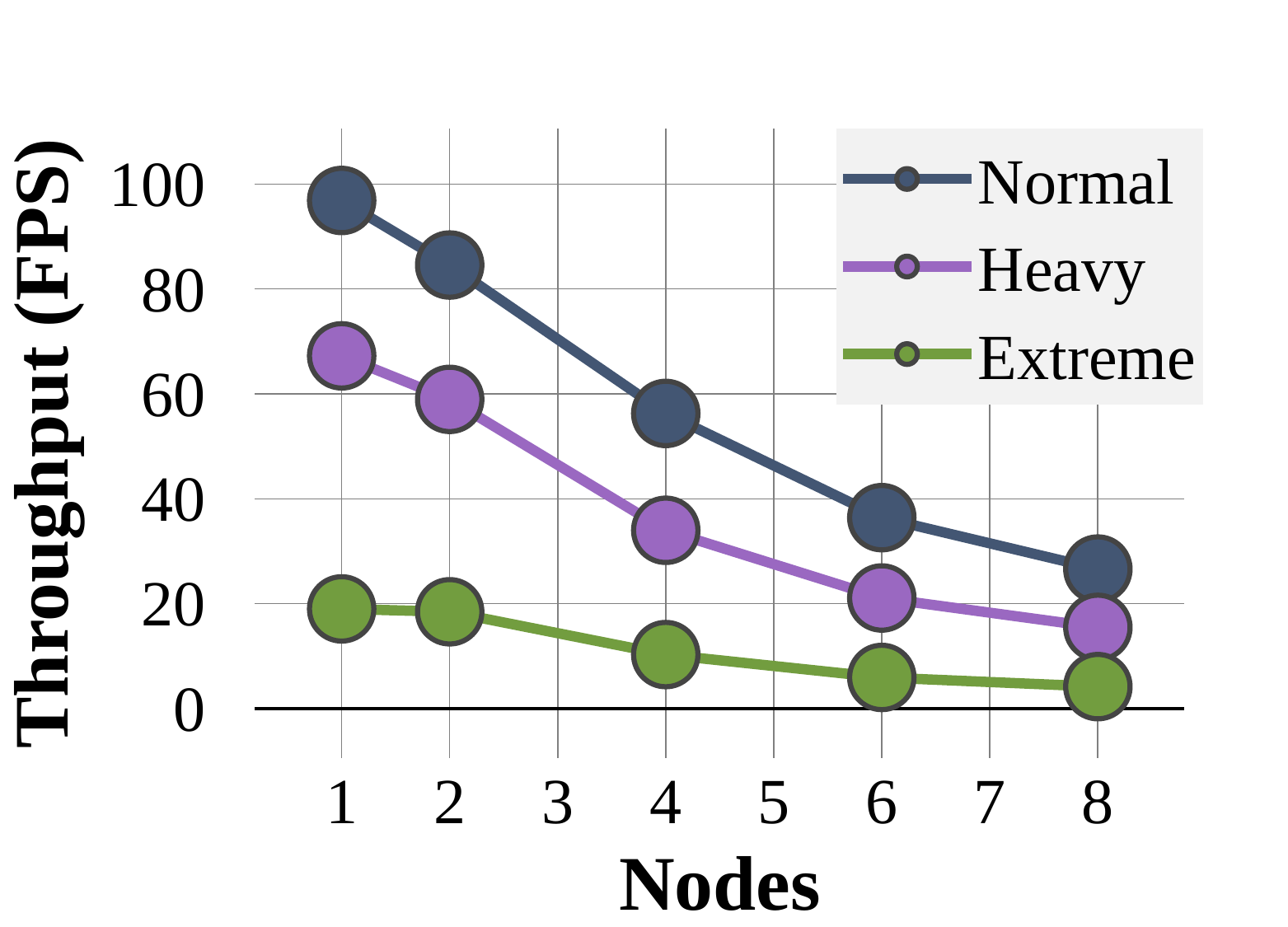}}\\
  \end{tabular}
  \caption{Throughput of Ancilia with respect to number of nodes across different crowd densities. Hardware details can be seen in \cref{tab:sys_configs}.}
  \label{fig:trend}
\end{figure*}

Another interesting observation is that CTR-GCN \cite{chen2021channel} actually performed noticeably better at half throughput than at full throughput. This means that CTR-GCN is more suited to taking advantage of the higher temporal window allowed when using half throughput. This is something to consider when choosing an action recognition model when a real-time throughput of 30 FPS cannot be guaranteed.


\subsubsection{High-level Task - Anomaly Detection} \label{sec:anomaly}

Using the ShanghaiTech dataset \cite{liu2018ano_pred} we train two state-of-the-art anomaly detection models, GEPC \cite{markovitz2020graph} and MPED-RNN \cite{morais2019learning}, using both data generated by Ancilia and the data provided by the original authors. The same training strategy from \cref{sec:action} is used, with both models trained in full (20 FPS) and half (10 FPS) modes. 
GEPC is trained for 25 epochs with a window size of 30 and stride of 20 using Adam optimizer with a learning rate of 1e-4, weight decay of 1e-5, and batch size of 512. 
MPED-RNN is trained with an input window size of 30, a reconstruction window of 12, and a prediction window of 6. The model is trained for 5 epochs using the Adam optimizer with a learning rate of $1e^{-3}$ and a batch size of 265.

\begin{table}[h]
\renewcommand{\arraystretch}{1.2}
\centering
\vspace{-5pt}
\caption{AUC ROC, AUC PR, and EER on ShanghaiTech dataset \cite{liu2018ano_pred} in full and half throughput modes for GEPC \cite{markovitz2020graph} and MPED-RNN \cite{morais2019learning}.}
\vspace{-6pt}
\label{tab:anomaly}
\begin{tabular}{c||c|c||ccc}
\rowcolor{DarkGray}
Model & Data & FPS & \multicolumn{1}{c|}{AUC ROC} & \multicolumn{1}{c|}{AUC PR} & \multicolumn{1}{l}{EER} \\ \hline \hline
\multirow{4}{*}{GEPC \cite{markovitz2020graph}} & \multirow{2}{*}{\cite{markovitz2020graph}} & 10 & 0.6906 & 0.5951 & 0.35 \\ \cline{3-6} 
 &  & 20 & 0.7372 & 0.6427 & 0.31 \\ \cline{2-6} 
 & \multirow{2}{*}{Ours} & 10 & 0.6888 & 0.5905 & 0.35 \\ \cline{3-6} 
 &  & 20 & 0.7223 & 0.6023 & 0.32 \\ \hline \hline
\multirow{4}{*}{MPED-RNN \cite{morais2019learning}} & \multirow{2}{*}{\cite{morais2019learning}} & 10 & 0.6645 & 0.5733 & 0.37 \\ \cline{3-6} 
 &  & 20 & 0.7023 & 0.5869 & 0.36 \\ \cline{2-6} 
 & \multirow{2}{*}{Ours} & 10 & 0.6685 & 0.5661 & 0.37 \\ \cline{3-6} 
 &  & 20 & 0.6679 & 0.5487 & 0.37
\end{tabular}
\vspace{-5pt}
\end{table}

The results of this experiment can be seen in \cref{tab:anomaly}. In line with current practices, we report Area Under the Receiver Operating Characteristic Cure (AUC ROC), Area Under the Precision-recall Curve (AUC PR), and the Equal Error Rate (EER). With GEPC, we can see that Ancilia more than measures up to the task, with only a 1.5\% drop in AUC ROC at full throughput and less than a 0.2\% drop in AUC ROC at half throughput. AUC PR shows a more substantial drop of 4\% at full throughput, but goes down to less than 0.5\% at half throughput. Equal Error Rates are almost identical, seeing almost no change (less than 0.01) when using Ancilia. MPED-RNN, which displayed lower overall accuracy in all regards to begin with, sees a more significant drop in AUC ROC at full throughput, losing 3.5\%. However, at half throughput the AUC ROC actually increases when using Ancilia, though only by 0.5\%. The AUC PR results mirror that of GEPC, dropping 3.8\% at full throughput and 0.7\% at half throughput. The Equal Error Rates are once again nearly identical. Being able to perform a high-level task such as anomaly detection while maintaining accuracies so close to current SotA in research, demonstrates Ancilia's ability to produce quality data, suitable for intelligent surveillance applications.

\subsection{Real-time System Performance}
\begin{table*}[t]
    \centering
    \vspace{-5pt}
    \caption{System configurations. Stats are per CPU/GPU of the listed type.}
    \vspace{-6pt}
    \label{tab:sys_configs}
    \begin{tabular}{c||c|c|c||c|c|c}
        \cellcolor{DarkGray} & \multicolumn{3}{d||}{Processor} & \multicolumn{3}{d}{GPU}  \\ 
        \cellcolor{DarkGray} \multirow{-2}{*}{Name} & \cellcolor{Gray}Model & \cellcolor{Gray}Cores & \cellcolor{Gray}Clock Speed & \cellcolor{Gray}Model & \cellcolor{Gray}CUDA Cores & \cellcolor{Gray}VRAM \\
        \hline \hline
        Server A & 2$\times$ EPYC 7513 & 32 & 2.6 GHz & 4$\times$ V100 & 5120 & 32 GB  \\
        Server B & 2$\times$ Xeon E5-2640 v4 & 10 & 2.4 GHz & 2$\times$ Titan V & 5120 & 12 GB   \\
        Workstation & Threadripper Pro 3975WX & 32 & 3.50 GHz & 3$\times$ A6000 & 10752 & 48 GB  \\
    \end{tabular}
    \vspace{-5pt}
\end{table*}

\begin{table*}[t]
    \centering
    \caption{Average throughput and latency. Data collected using the Workstation with varying local node counts.}
    \vspace{-6pt}
    \begin{tabular}{c|c||C{1.5cm}c|C{1.5cm}c|C{1.5cm}c}
       \rowcolor{DarkGray} &  & \multicolumn{2}{c|}{Server A} & \multicolumn{2}{c|}{Server B} & \multicolumn{2}{c}{Workstation} \\ \cline{3-8}
       \rowcolor{DarkGray} \multirow{-2}{*}{\shortstack{Crowd Density}} & \multirow{-2}{*}{Nodes} &  \cellcolor{Gray} FPS & \cellcolor{Gray} Latency (s) & \cellcolor{Gray} FPS & \cellcolor{Gray} Latency (s) & \cellcolor{Gray} FPS & \cellcolor{Gray} Latency (s) \\ \hline \hline
        \multirow{5}{*}{\shortstack{Normal \\ \\ \\ (70 detections \\ per second)}} & 1 & 82.31 & 1.17 & 52.45 & 1.52 & 96.88 & 0.87 \\
         & 2 & 77.59 & 1.15 & 39.50 & 2.05 & 84.57 & 1.00 \\
         & 4 & 53.40 & 1.60 & - & - & 56.27 & 1.58 \\
         & 6 & 33.43 & 1.99 & - & - & 36.40 & 2.27 \\
         & 8 & 23.64 & 2.05 & - & - & 26.60 & 2.84 \\ \hline
        \multirow{5}{*}{\shortstack{Heavy \\ \\ \\ (216 detections \\ per second)}} & 1 & 57.47 & 1.80 & 38.97 & 2.62 & 67.25 & 1.53 \\
         & 2 & 50.05 & 2.07 & 32.85 & 3.16 & 58.95 & 1.77 \\
         & 4 & 32.09 & 3.45 & - & - & 33.99 & 3.98 \\
         & 6 & 18.51 & 4.17 & - & - & 21.08 & 6.89 \\
         & 8 & 13.35 & 5.87 & - & - & 15.48 & 9.54 \\ \hline
        \multirow{5}{*}{\shortstack{Extreme \\ \\ \\ (744 detections \\ per second)}} & 1 & 19.84 & 5.29 & 14.67 & 7.37 & 19.00 & 5.73 \\
         & 2 & 20.76 & 5.09 & 16.56 & 6.54 & 18.45 & 5.81 \\
         & 4 & 10.95 & 11.64 & - & - & 10.29 & 12.49 \\
         & 6 & 6.25 & 20.70 & - & - & 5.93 & 21.87 \\
         & 8 & 4.53 & 28.48 & - & - & 4.18 & 31.19 \\
    \end{tabular}
    \label{tab:sys_results}
    \vspace{-5pt}
\end{table*}

Algorithmic accuracy is vital for ensuring the information provided by high-level cognitive tasks is beneficial for surveillance applications. However, Ancilia's ability to perform in real-time is equally important. We conduct a series of experiments, evaluating the runtime performance of Ancilia on different hardware, with different scenario intensities, and for increasing number of local nodes per hardware device. We focus on the performance of the local node, as the global node is completely decoupled from the algorithmic pipeline and has no noticeable effect on throughput or latency.

\begin{figure}
    \includegraphics[trim={5 0 40 55},clip,width=0.9\linewidth]{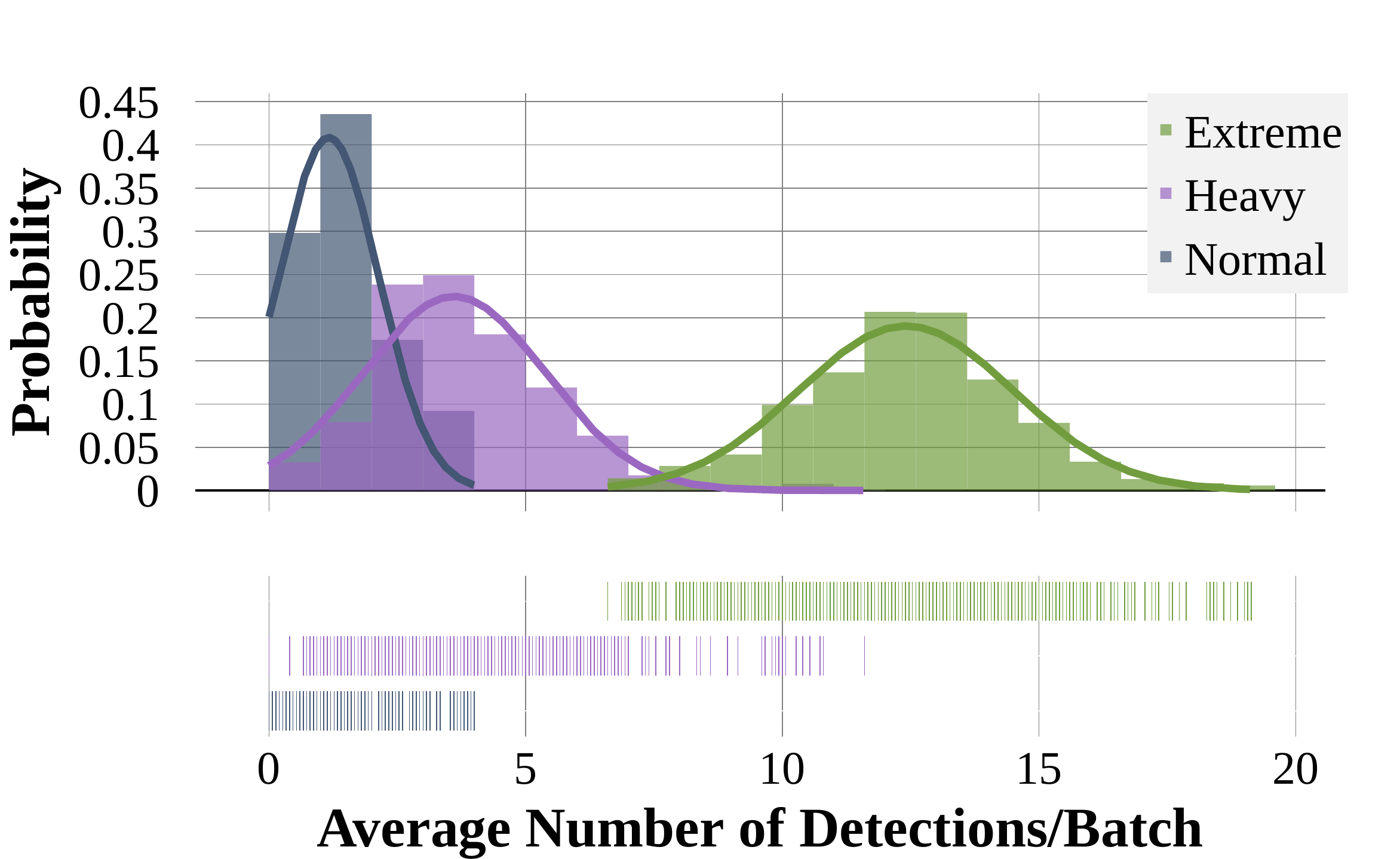} \\
    \includegraphics[trim={5 0 40 55},clip,width=0.9\linewidth]{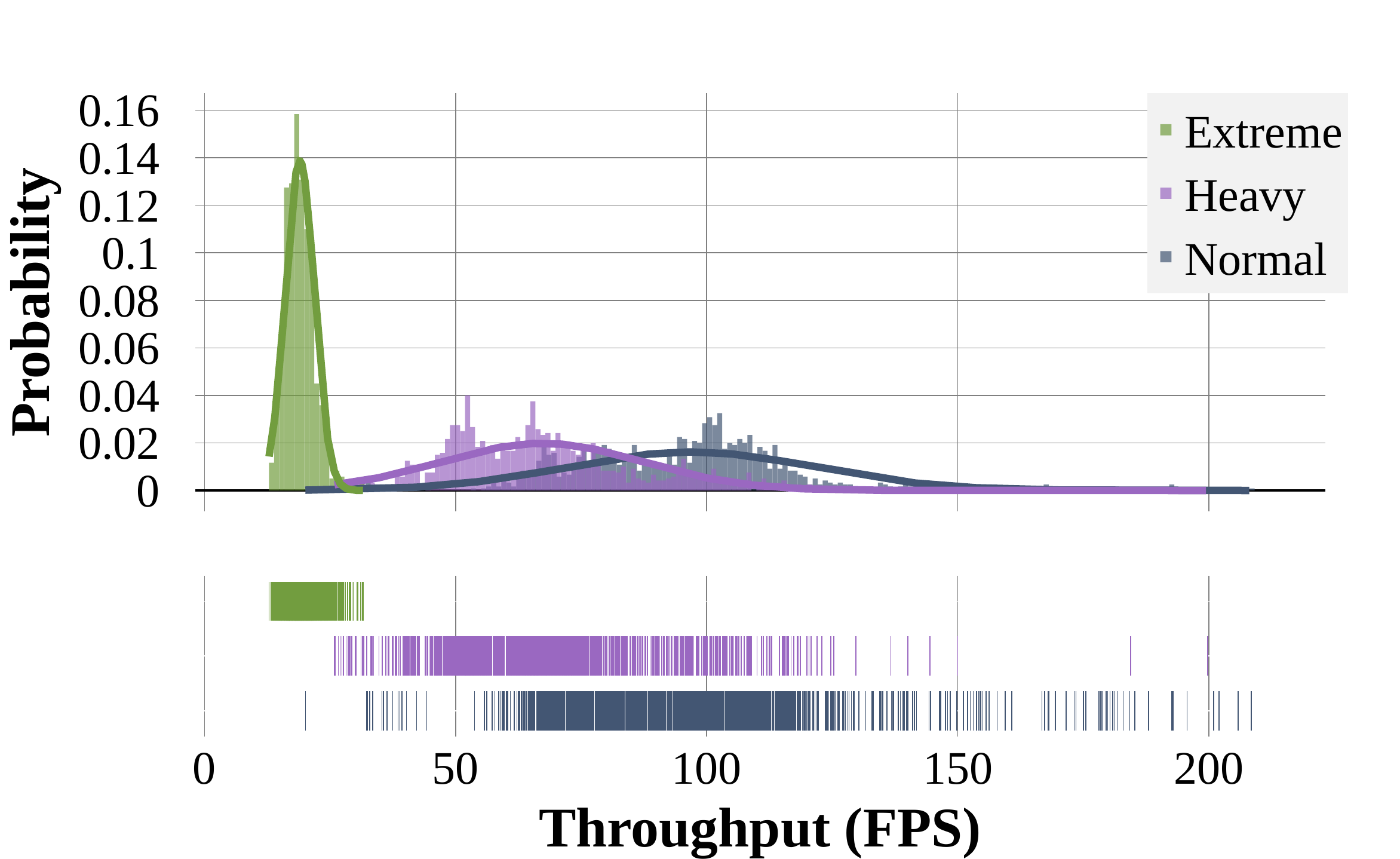} \\
  \vspace{-15pt}
  \caption{Distribution of detections for different crowd densities and its effect on throughput. Data collected using the Workstation with a single local node.}
  \label{fig:stability}
  \vspace*{-\baselineskip}
\end{figure}

We choose three different hardware configurations for these experiments: a high-end server, a lower-end server, and a high-end workstation, as seen in \cref{tab:sys_configs}. For our scenarios, we use the DukeMTMC-video dataset \cite{DukeMTMC} and pick three scenes with different crowd densities: normal density, heavy density, and extreme density. The distribution of detection density in each scenario, as well as their effect on throughput, can be seen in \cref{fig:stability}. Note that what is considered "normal density" will change based on application environment, which is why we report on such a wide range. Each video lasts for 32k frames, with 7k frames warm-up and cool-down. We test using 1, 2, 4, 6, and 8 local nodes on a single system, showing how throughput and latency scale in such cases. Each experiment is conducted three times, the throughput and latency averaged across runs. The results of these experiments can be seen in \cref{tab:sys_results} and \cref{fig:trend}.

Under normal crowd density, Server A and Workstation are both able to achieve over 50 FPS with up to four local nodes, with an end-to-end latency of 1.60 and 1.58 seconds respectively. This is well above FPS required by action recognition and anomaly detection algorithms at full throughput, and the latency is low enough to be suitable for most surveillance applications where the main concern is to notify authorities in time for appropriate response. Both Server A and Workstation are able to handle 6 local nodes in the normal scenario while maintaining above 30 FPS. Workstation is able to maintain above 26 FPS while running all 8 local nodes, while Server A drops to just below 24 FPS at 8 local nodes. Server B is able to achieve over 50 FPS with a single node but falls just short of 40 FPS while handling two nodes simultaneously. Due to having only two GPUs with limited VRAM, Server B was unable to run 4 or more nodes concurrently.

Heavy crowd density proves more challenging, with both Server A and Workstation only able to achieve above 30 FPS with up to 4 nodes. The end-to-end latency is also longer than it was under normal crowd density, with all systems seeing between a 50\% to 100\% increase in most cases, and up to a 230\% increase at the mose extreme. Server A and Workstation are able to mainatin above 15 FPS at 6 and 8 nodes respectively, while Server A drops to just above 13 FPS at 8 nodes. Server B behaves similarly to how it did with normal crowd density, still able to maintain above 30 FPS for up to 2 nodes, though with slightly low throughput. Assuming only half throughput was needed for high-level tasks, Server B would still be suitable for running up to two nodes.



With the extreme crowd density scenario, Ancilia begins to struggle. None of the systems are able to achieve above 30 FPS even with a single camera, putting full throughput action recognition out of reach. Server A is able to achieve above 20 FPS with 2 nodes (but notably not with 1) and Workstation fall short even with 1 node. Both Server A and Workstation can maintain above 10 FPS at 4 nodes, but both drop to around 6 and 4 FPS at 6 and 8 nodes, respectively. \cite{ActionSurvey} argues that 5 FPS is suitable for tacking pedestrians, and while that is true, high-level tasks that rely on detailed human motion, such as action recognition and anomaly detection, often struggle for accuracy when running below 10 FPS. Another issue is with the increased latency. Running 6+ nodes, Server A and Workstation have latencies over 20 seconds, which is suitable for many surveillance applications, but might be too much for those that require sharper response times. Combined with the low throughput, it becomes difficult to recommend running more than 4 nodes on a single system with Ancilia when operating under extreme crowd density, expect for applications where low throughput and high latency are not as much of a concern. Server B is unable to achieve 30 FPS, but does stay around 15 FPS for both 1 and 2 nodes, making it suitable for half throughput in action recognition and anomaly detection.

Interestingly, with extreme crowd density we start to see unusual behavior with both Servers having worse performance with a singe node than they do with 2 nodes. This is likely caused by the abundance of CPU resources available to them with their dual CPU configuration and a single node being unable to fully utilize them. As such, the behavior of both servers in the extreme crowd density scenario does not start to match the expected behavior and mimic the other systems until multiple nodes are being run simultaneously. This behavior is not too concerning, considering it does not make sense to purchase such a high-end server class machine for only running a single local node, when a more latency focused workstation would be both cheaper and more effective.

\begin{figure}
  \centering
    \includegraphics[trim={10 5 65 75},clip,width=0.6\linewidth]{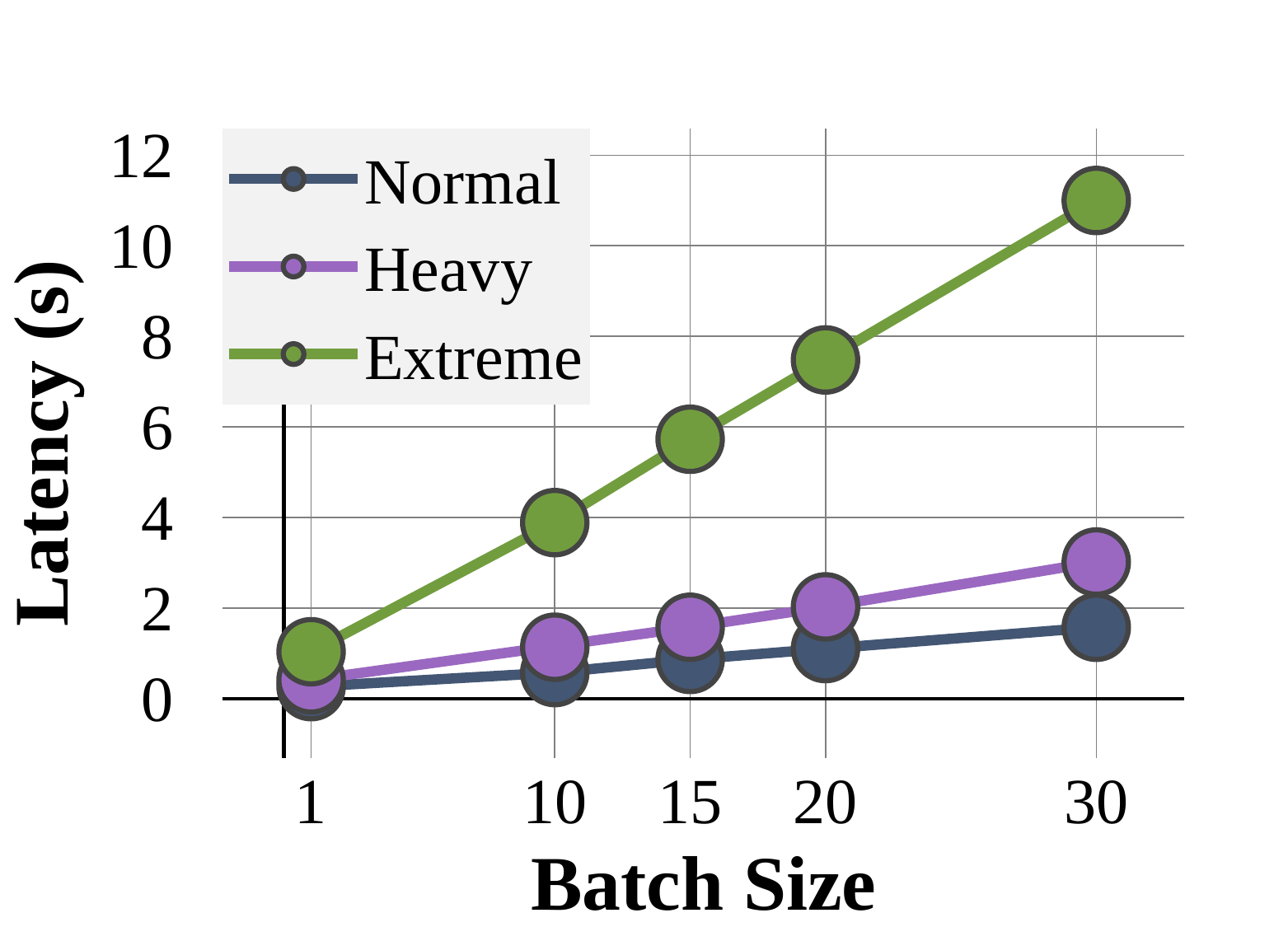} \\
    \vspace{5pt}
    \includegraphics[trim={10 5 65 75},clip,width=0.6\linewidth]{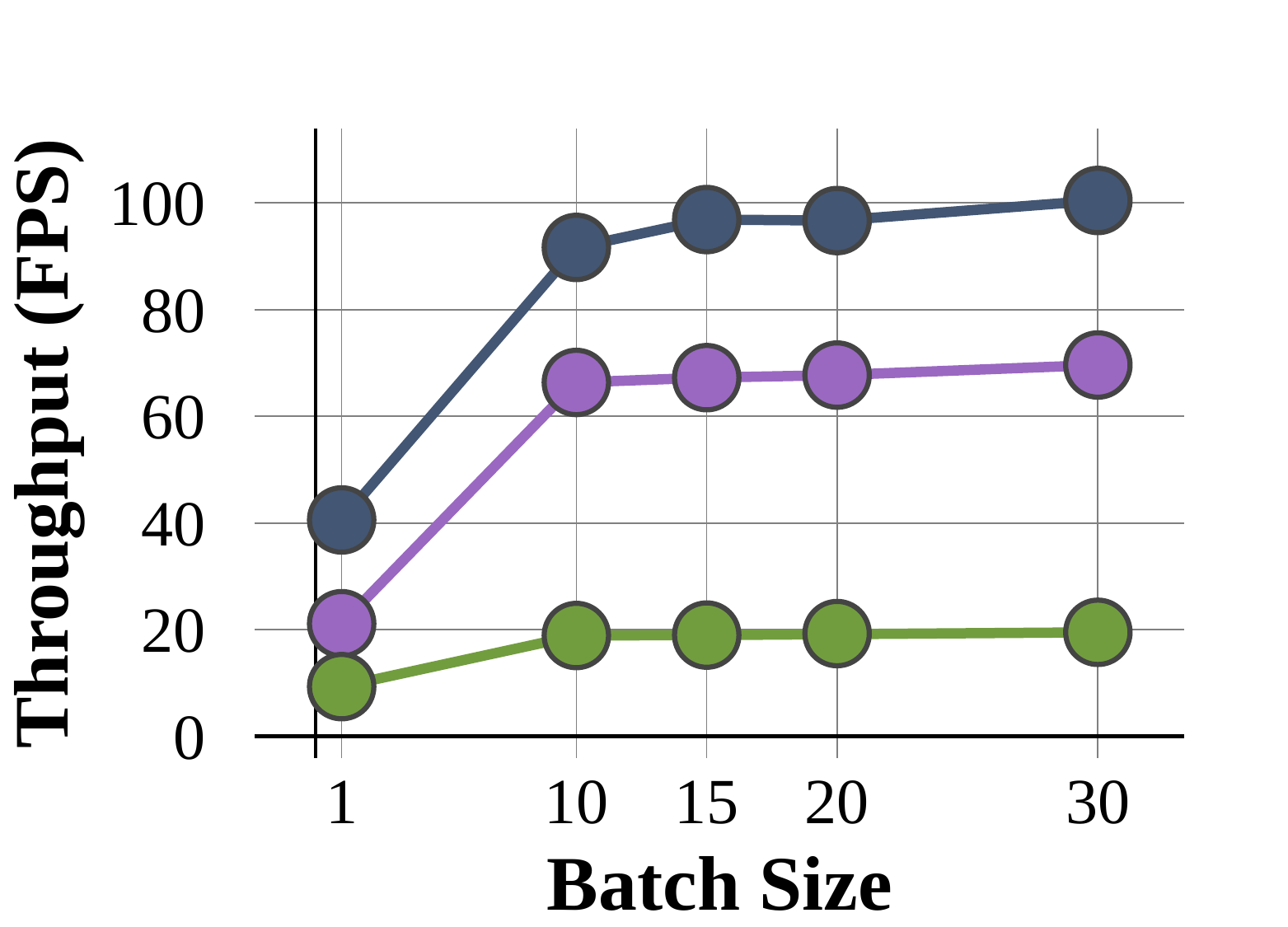} \\
    
  \caption{Throughput and latency trends with respect to batch size across different crowd densities. Data collected using Workstation with a single local node.}
  \label{fig:batchtrend}
  \vspace*{-\baselineskip}
\end{figure}

Overall, Ancilia is able to meet the needs of high-level cognitive tasks while still achieving performance suitable for real-time intelligent surveillance applications. Exact performance is dependent on both the hardware used and the intensity of the scene, but these results show that even for the most extreme of scenarios, Ancilia can be used to provide intelligent assistance to surveillance applications.

\begin{table}[]
\centering
\caption{Effect of different batch sizes on throughput and latency.}
\vspace{-6pt}
\label{tab:batch}
\begin{tabular}{c||c|cc}
\rowcolor{DarkGray}
Crowd Density            & Batch Size & FPS    & Latency (s)\\ \hline \hline
\multirow{5}{*}{Normal}  & 1          & 40.58  & 0.27              \\
                         & 10         & 91.66  & 0.58          \\
                         & 15         & 96.88  & 0.87          \\
                         & 20         & 96.69  & 1.11          \\
                         & 30         & 100.46 & 1.58          \\ \hline
\multirow{5}{*}{Heavy}   & 1          & 21.13  & 0.42              \\
                         & 10         & 66.35  & 1.14          \\
                         & 15         & 67.25  & 1.58          \\
                         & 20         & 67.73  & 2.03          \\
                         & 30         & 69.62  & 3.02          \\ \hline
\multirow{5}{*}{Extreme} & 1          & 9.34   & 1.04              \\
                         & 10         & 18.92  & 3.89          \\
                         & 15         & 19.00  & 5.73          \\
                         & 20         & 19.28  & 7.48          \\
                         & 30         & 19.52  & 11.00        
\end{tabular}
\end{table}

\subsection{Effect of Batch Size on Real-time Performance} \label{sec:batch_size}

To understand the effect of batch size on end-to-end latency and throughput, we test using a single node on Workstation but varying the batch size. The results of this can be seen in \cref{fig:batchtrend}. As expected, both latency and throughput increase with batch size across all densities. The jump in throughput from a batch size of 1 to a batch size of 10 is the most dramatic, with diminishing returns using larger batch sizes, while increases in latency tend to be more proportional. However, due to the high-level tasks needing 30 frames, the end-to-end latency is not directly representative of the latency of performing high-level tasks.

Overall, a balance needs to be struck between throughput, end-to-end latency, and batch size. Too high of an end-to-end latency will effect the speed at which detected objects of interest raise alarms, while a lower throughput can affect high-level task accuracy, as seen in \cref{sec:highlevel}. Likewise, having too small of a batch size means more batches need to be processed before high-level tasks can operate. A batch size of 15 strikes this balance well, with less than a second of end-to-end latency of 0.87 seconds and a throughput of 96.88 FPS in normal density, and only needing to process two batches for high-level tasks. This proves similar for heavy and extreme crowd densities as well, though the throughput is higher and latency is lower, as expected. 

\section{Conclusion}\label{sec:Conclusion}

In this article we presented Ancilia, an end-to-end scalable intelligent video surveillance system for the Artificial Intelligence of Things. Through empirical evaluation, Ancilia has demonstrated its ability to bring state-of-the-art artificial intelligence to real-world surveillance applications. Ancilia performs high-level cognitive tasks (i.e. action recognition and anomaly detection) in real-time, all while respecting ethical and privacy concerns common to surveillance applications.

\section*{Acknowledgments}

This research is supported by the National Science Foundation (NSF) under Award No. 1831795 and NSF Graduate Research Fellowship Award No. 1848727.

\bibliographystyle{IEEEtran}
\scriptsize
\bibliography{references}

\newpage

\section*{Biography}
\vspace{-35pt}
\begin{IEEEbiography}[{\includegraphics[width=1in,height=1.12in,keepaspectratio]{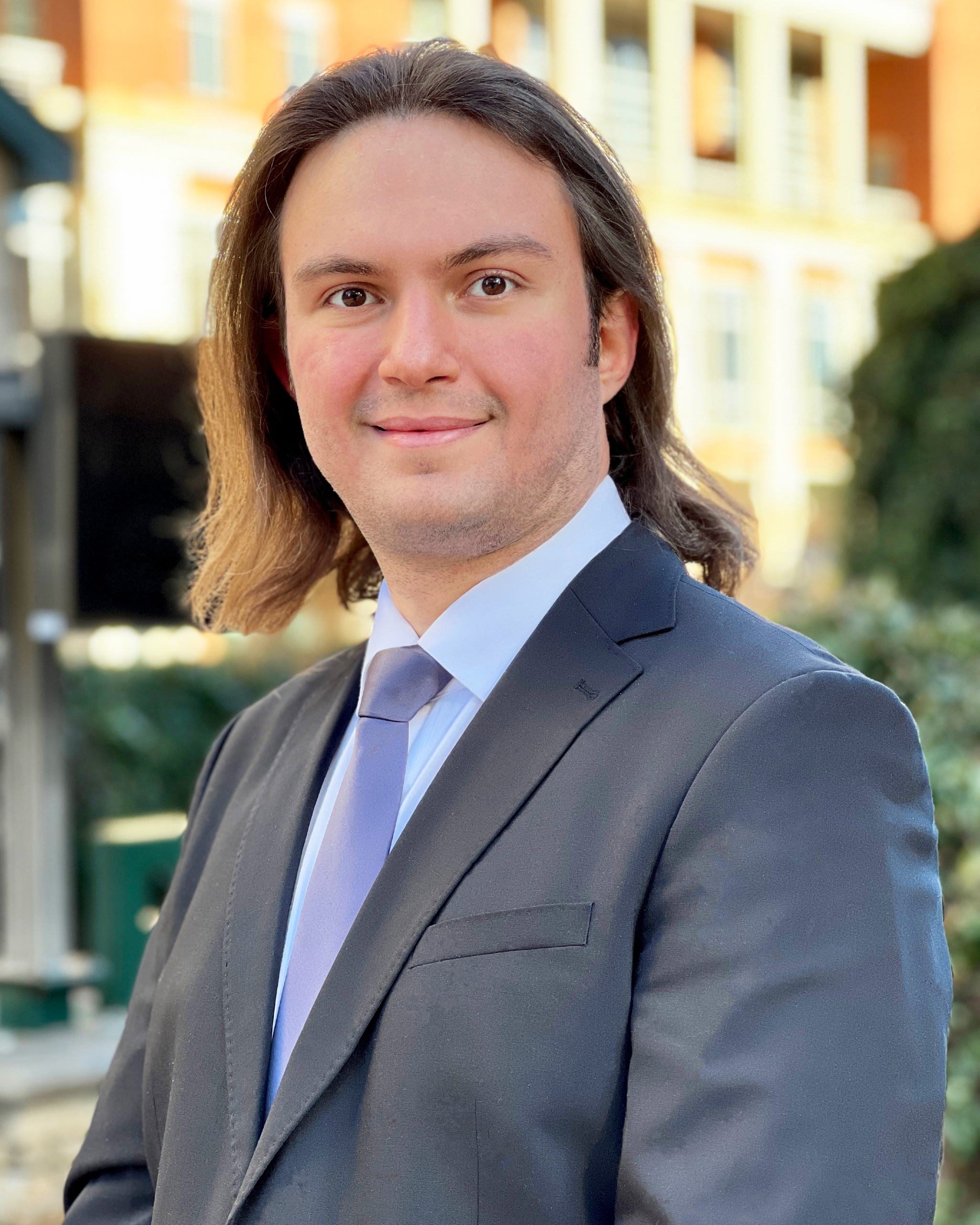}}]{Armin Danesh Pazho} (S’22) is currently a Ph.D. student at the University of North Carolina at Charlotte, NC, United States. With a focus on Artificial Intelligence, Computer Vision, and Deep Learning, his research delves into the realm of developing AI for practical, real-world applications and addressing the challenges and requirements inherent in these fields. Specifically, his research covers action recognition, anomaly detection, person re-identification, human pose estimation, and path prediction.
\end{IEEEbiography}

\begin{IEEEbiography}[{\includegraphics[width=1in,height=1.25in,keepaspectratio]{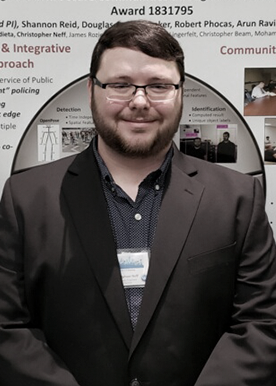}}]{Christopher Neff}
(S’18) 
is a National Science Foundation Graduate Research Fellow and Doctoral Candidate at the University of North Carolina at Charlotte. His dissertation focus is on tackling the challenges of bringing human-centric computer vision to real-world applications. His previous work focuses on person re-identification, human pose estimation, action recognition, real-time system development, lightweight algorithms, noisy data, domain shift, and real-world applications.
\end{IEEEbiography}

\begin{IEEEbiography}[{\includegraphics[width=1in,height=1.1in,keepaspectratio]{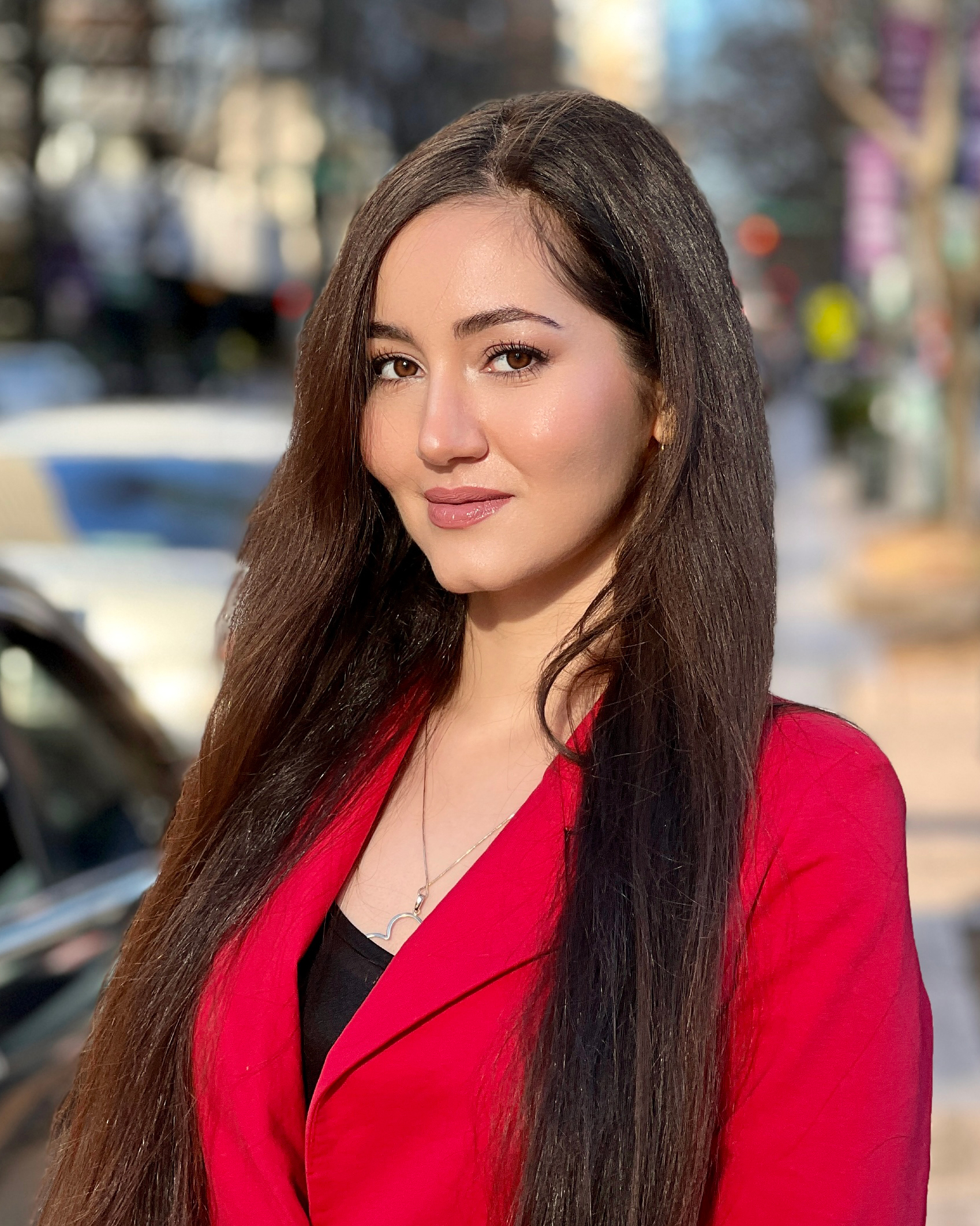}}]{Ghazal Alinezhad Noghre} (S’22) is currently pursuing her Ph.D. in Electrical and Computer Engineering at the University of North Carolina at Charlotte, NC, United States. Her research concentrates on Artificial Intelligence, Machine Learning, and Computer Vision. She is particularly interested in the applications of anomaly detection, action recognition, and path prediction in real-world environments, and the challenges associated with these fields.
\end{IEEEbiography}

\begin{IEEEbiography}[{\includegraphics[width=1in,height=1.1in,keepaspectratio]{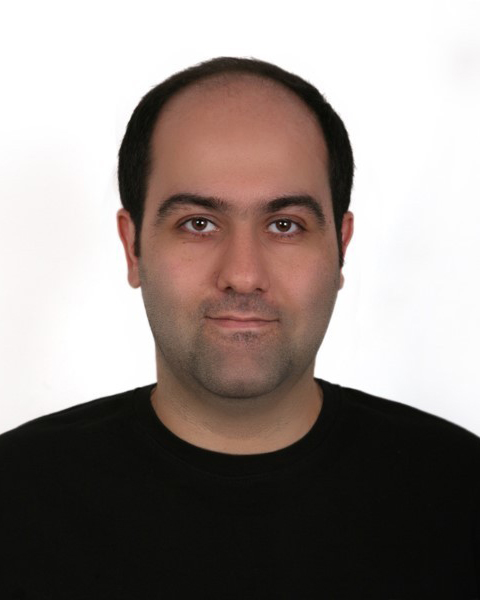}}]{Babak Rahimi Ardabili} is a Ph.D. student in the Public Policy Analysis program at the University of North Carolina at Charlotte, United States. His main research area is emerging technologies policy making. He mainly focuses on the intersection of Artificial Intelligence and policy from a privacy perspective and the challenges of bringing the technology to the community.  
\end{IEEEbiography}

\begin{IEEEbiography}[{\includegraphics[width=1in,height=1.125in,keepaspectratio]{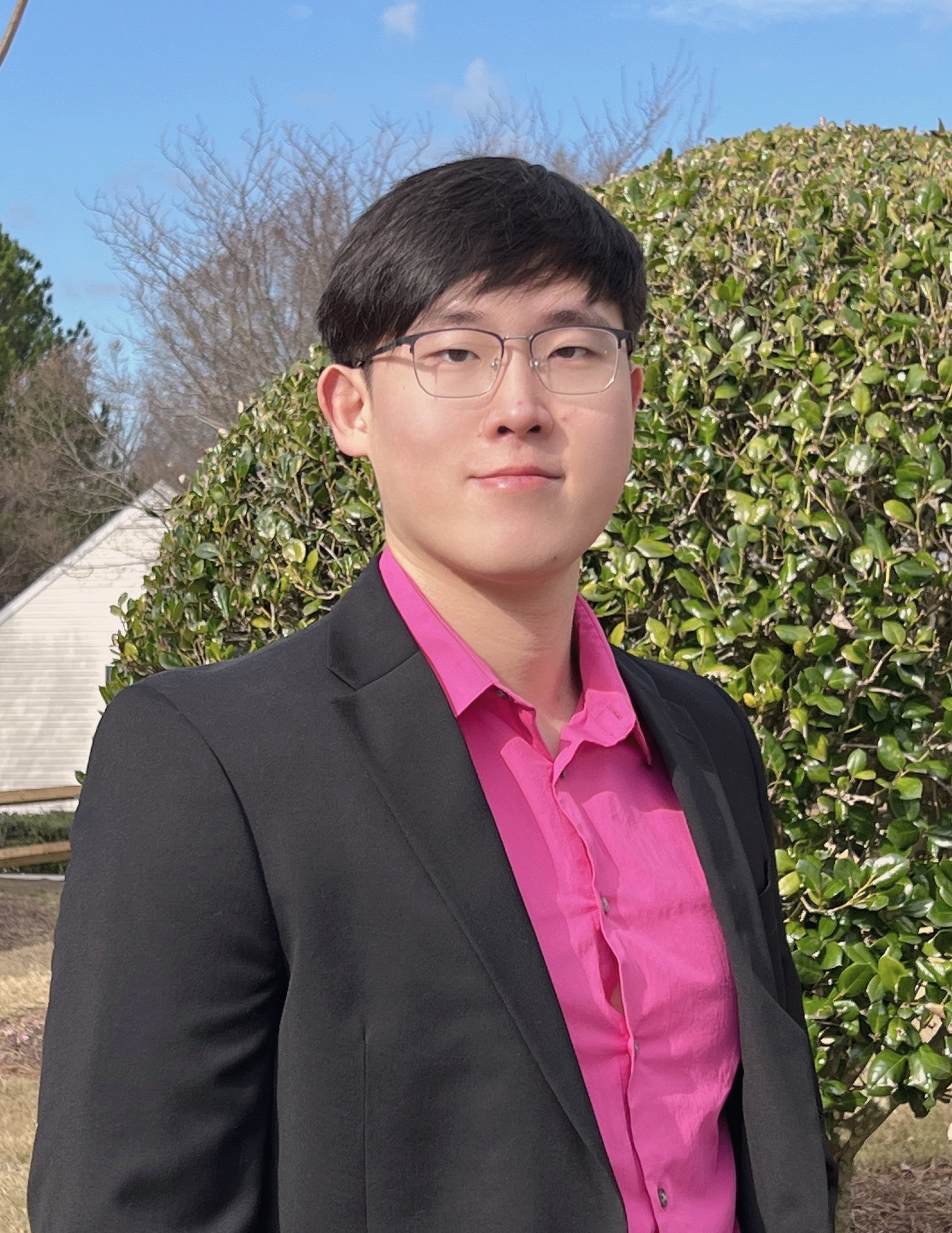}}]{Shanle Yao} is an Electrical  Engineering Graduate student from the University of North Carolina at Charlotte. His dissertation focus is on optimization and application of Computer Vision pipeline performance and throughput. His areas of interest include object detection, multiple objects tracking, human pose estimation, semantic segmentation and real-world applications.
\end{IEEEbiography}

\begin{IEEEbiography}
[{\includegraphics[width=1in,height=1.19in,keepaspectratio]{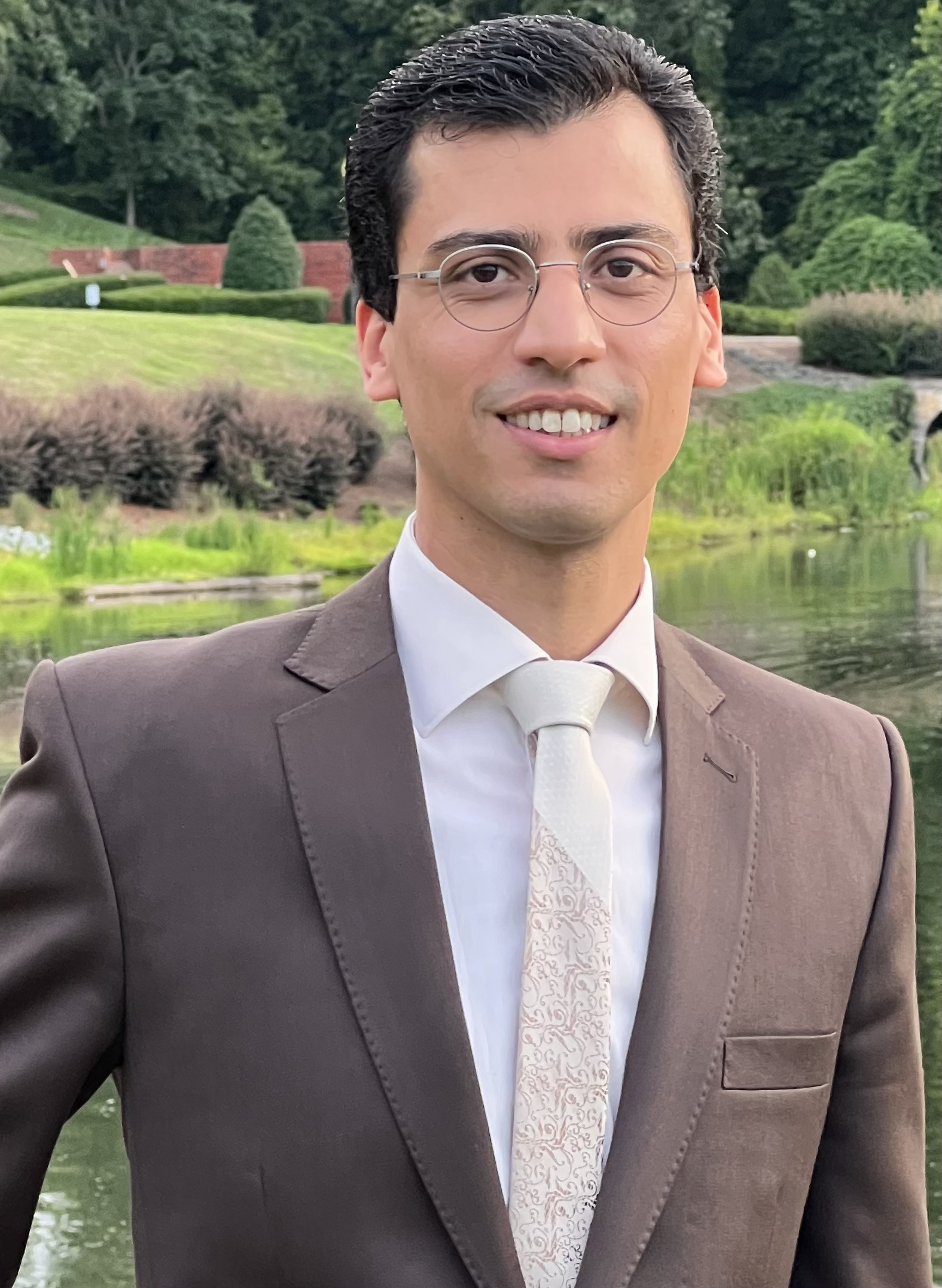}}]{Mohammedreza Baharani} is an ML researcher and edge system deployment engineer at ForesightCares. He received his Ph.D. in computer engineering in 2021 from the University of North Carolina at Charlotte, USA, and was a postdoctoral researcher at the TeCSAR Lab. His research focuses on the intersection of computer architecture engineering and machine learning, with the goal of enabling AI algorithms on edge devices to have a positive impact in fields such as healthcare.
\end{IEEEbiography}

\begin{IEEEbiography}[{\includegraphics[width=1in,height=1.25in,keepaspectratio]{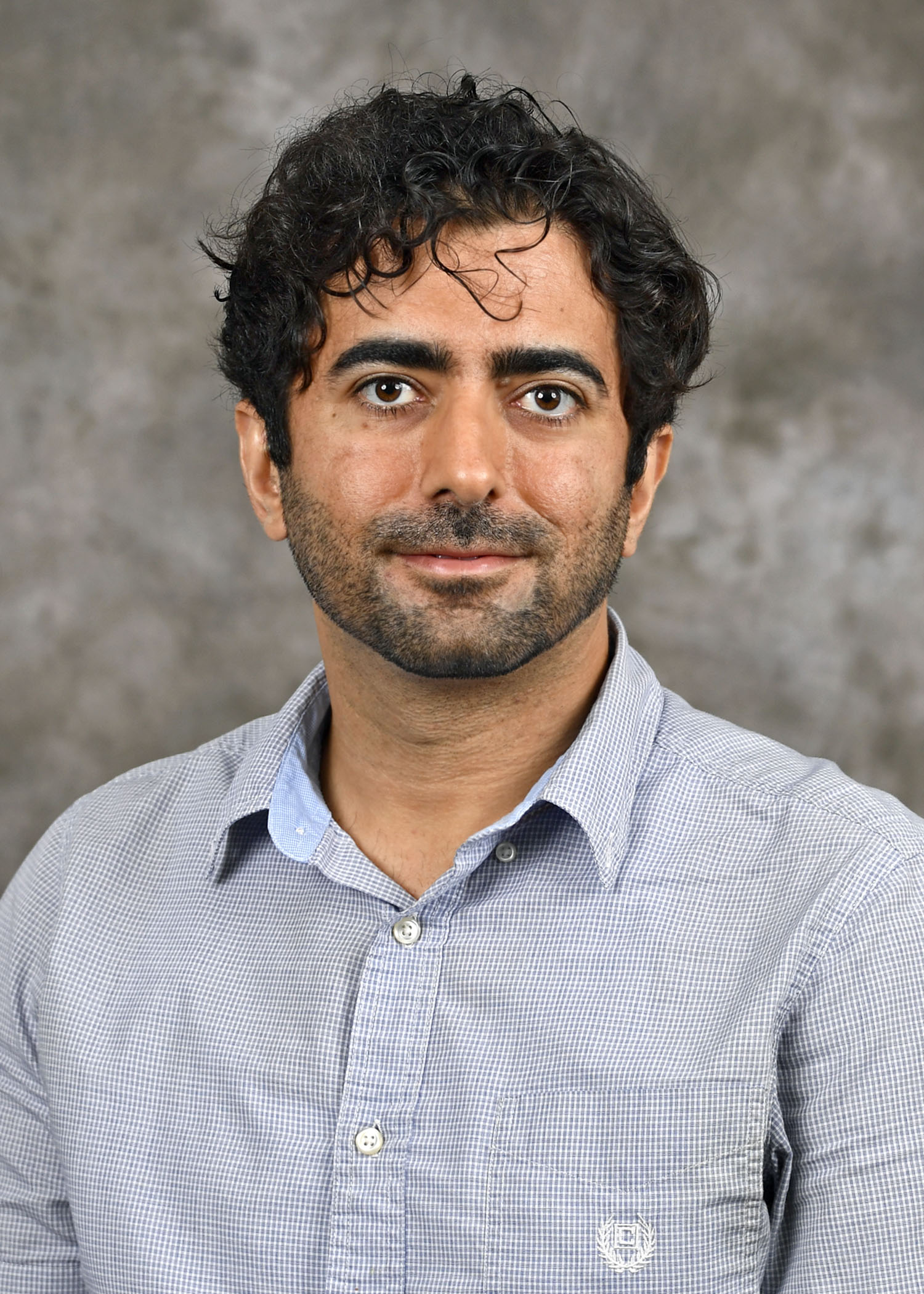}}]{Hamed Tabkhi}
(S’07–M’14)
is an Associate Professor in the Department of Electrical and Computer Engineering, University of North Carolina at Charlotte, USA.
He was a post-doctoral research associate at Northeastern University. Hamed Tabkhi received his Ph.D. degree in 2014 from Northeastern University under the direction of Prof. Gunar Schirner. Overall, his research focuses on transformative computer systems and architecture for cyber-physical, real-time streaming and emerging machine learning applications.
\end{IEEEbiography}

 




\vfill

\end{document}